\documentclass[acmsmall,nonacm]{acmart}

\AtBeginDocument{%
  }

\setcopyright{acmlicensed}
\copyrightyear{2018}
\acmYear{2024}
\acmDOI{XXXXXXX.XXXXXXX}

\acmJournal{TIST}
\acmVolume{15}
\acmNumber{4}
\acmArticle{1}
\acmMonth{4}

\usepackage{times}
\usepackage{soul}
\usepackage{url}
\usepackage{amsmath}
\usepackage{amsthm}
\usepackage{booktabs}
\usepackage{algorithm}
\usepackage{algorithmic}
\usepackage[switch]{lineno}
\usepackage{subcaption}
\usepackage{xspace}
\usepackage{xcolor}
\usepackage{diagbox}  
\usepackage{enumitem}
\usepackage{multirow}

\usepackage{caption}
\usepackage{subcaption}

\newcommand{\modified}[1]{\color{black}{#1}} 
\newcommand{\modifiedblue}[1]{\color{black}{#1}} 

\newcommand{\ourf}[1]{{\textsc{#1}}}
\newcommand{\gptt}{\ourf{GPT-3.5}\xspace}
\newcommand{\gptf}{\ourf{GPT-4}\xspace}
\newcommand{\pythia}{\ourf{Pythia-12B}\xspace}
\newcommand{\gemini}{\ourf{Gemini-1.5F}\xspace}

\usepackage{natbib}

\makeatletter
\newcommand\footnoteref[1]{\protected@xdef\@thefnmark{\ref{#1}}\@footnotemark}
\makeatother

\begin{document}

\title{Do LLMs Dream of Ontologies?}


\author{Marco Bombieri}
\email{marco.bombieri_01@univr.it}
\orcid{0000-0002-8607-8495}
\authornotemark[1]
\affiliation{%
  \institution{University of Verona}
  \city{Verona}
  \country{Italy}
}

\author{Paolo Fiorini}
\email{paolo.fiorini@univr.it}
\affiliation{%
  \institution{University of Verona}
  \city{Verona}
  \country{Italy}
}

\author{Simone Paolo Ponzetto}
\affiliation{%
  \institution{University of Mannheim}
  \city{Mannheim}
  \country{Germany}}
\email{ponzetto@uni-mannheim.de}

\author{Marco Rospocher}
\affiliation{%
  \institution{University of Verona}
  \city{Verona}
  \country{Italy}
}
\email{marco.rospocher@univr.it}

\renewcommand{\shortauthors}{Bombieri et al.}

\begin{abstract}
{\modifiedblue{%
Large Language Models (LLMs) have demonstrated remarkable performance across diverse natural language processing tasks, yet their ability to memorize structured knowledge remains underexplored. In this paper, we investigate the extent to which general-purpose pre-trained LLMs retain and correctly reproduce concept identifier (ID)–label associations from publicly available ontologies. We conduct a systematic evaluation across multiple ontological resources, including the Gene Ontology, Uberon, Wikidata, and ICD-10, using LLMs such as \pythia, \ourf{Gemini-1.5-Flash}, \gptt, and \gptf. Our findings reveal that only a small fraction of ontological concepts is accurately memorized, with \gptf demonstrating the highest performance.
To understand why certain concepts are memorized more effectively than others, we analyze the relationship between memorization accuracy and concept popularity on the Web. Our results indicate a strong correlation between the frequency of a concept's occurrence online and the likelihood of accurately retrieving its ID from the label. This suggests that LLMs primarily acquire such knowledge through indirect textual exposure rather than directly from structured ontological resources. Furthermore, we introduce new metrics to quantify prediction invariance, demonstrating that the stability of model responses across variations in prompt language and temperature settings can serve as a proxy for estimating memorization robustness.
}}
\end{abstract}





\maketitle

\section{Introduction}

Large Language Models (LLMs) have revolutionized how natural language is computationally processed. Pre-trained models, such as {\modifiedblue{GPT-3~\citep{gpt3}}}, LLaMA~\citep{touvron2023llama}, and BLOOM~\citep{workshop2023bloom} have shown that human-level performance can be achieved on various tasks {\modified{(e.g., those found in MMLU~\citep{hendryckstest2021} or ChatBot Arena~\cite{chiang2024chatbotarenaopenplatform}) with no or minimal task-specific adaptation.}} 

Recent works \cite[\emph{inter alia}]{bib-unintended-memorization-2019,Bamman2023,Carlini2023,Zanzotto2023,bib-black-box-llm-2024,bib-pretraining-detection-2024,bib-entity-level-memorization-2024} have investigated memorization aspects related to these models, i.e., to what extent can the content used to train an LLM be extracted as-is through appropriate interaction with the model. These works have mainly focused on extracting the actual sequences of tokens used to train the model. 
{\modified{Here, we investigate instead a different aspect of memorization, namely to what extent general-purpose pre-trained LLMs can memorize \textit{information}, meaning the ability of an LLM to recall the association between two tokens that may not occur consecutively in the text, following an approach similar to \cite{bib-entity-level-memorization-2024}.}}\footnote{{\modifiedblue{We emphasize that this paper focuses on \emph{memorization} rather than \emph{learning}, acknowledging that the distinction between the two is complex and widely debated: while learning in some cases involves some degree of memorization, they remain distinct concepts \cite{memorization-learning, memorization-learning2}.}}}
To this end, we assess whether and to what extent some popular pre-trained general-purpose LLMs have memorized information from known ontologies, focusing on a basic and fundamental piece of information, namely the association of concept identifiers with their corresponding natural language labels (e.g., \texttt{GO:0001822} with \textit{kidney development}).

We first evaluate how much LLMs such as {\modifiedblue{\gptt, \gptf}}, {\modified{GEMINI-1.5-Flash}} and \pythia~\citep{Pythia2023} have acquired information about concepts from the text and can retrieve it when prompted without further training. Our experiments, {\modified{conducted on both domain-specific (Gene~\citep{geneontology}, Uberon~\citep{uberon} and ICD-10~\citep{ICD-10}) and general-purpose (Wikidata~\citep{wikidata}) ontologies}}, demonstrate that, while LLMs show awareness of each of the ontologies we consider, they have not memorized completely and consistently all concepts. Consequently, we next investigate which ontological concepts have been better memorized and why. Experimental findings indicate that the correct memorization of concepts varies among them while correlating with the frequency of their occurrence on the Web, thus suggesting that the information was not acquired (or at least not exclusively) by processing the ontology itself but rather from other textual Web materials mentioning it. Indeed, our experiments advocate that the more information on some ontology concepts is present in textual form on the Web, which is probably largely used as training material for LLMs, the better they memorize it. {\modified{Our findings may loosely resemble some theories about how human memory functions}}, such as the \textit{law of repetition}: indeed, while for humans, ``the speed and accuracy of remembering improve with repeated opportunities to study and retrieve the to-be-remembered material''~\citep{kahana_diamond_aka_2022}, it seems that also for LLMs, memorizing information about some ontology appears to depend on the number of times that information has been seen in the training material. We finally focus on quantifying the extent of memorization observed in previous experiments. For this, we ask for the very same information multiple times to the LLM, using different prompt repetitions, setting different temperature levels to control the model's output, and using different prompting languages. The experiments show that the more the information is popular on the Web and thus likely to be more often seen in the training material, the more the output of the model is consistent and accurate independently of the prompt language and temperature level used, thus suggesting that the invariance of the model output can be used as evidence of correct information memorization.
To foster further research on these aspects, we make all code and data publicly available.\footnote{\url{https://github.com/marcobombieri/do-LLM-dream-of-ontologies}} 

\section{Related work}
\label{sub:related}

Several recent works have studied the memorization of different kinds of information in language models for various purposes.
\citet{Zanzotto2023} investigate the interplay between memorization and performance in downstream tasks with BERT. By defining a measure for evaluating memorization from pre-training, they establish a correlation between highly memorized examples and improved classification. \citet{Bamman2023} explore memorization in ChatGPT and GPT-4 through a `data archaeology' approach. By employing a name cloze membership inference query, they show that these models have memorized a diverse range of copyrighted materials, with the extent of memorization linked to the prevalence of passages on the Web.  
\citet{Carlini2023} delve into the undesirable consequences of memorization in LLMs: they demonstrate that LLMs, when prompted appropriately, emit verbatim memorized training data, leading to privacy violations, degraded utility, and fairness issues, a problem already identified for clinical notes in BERT \citep{lehman-etal-2021-bert} and personal information in generative LLMs \cite{huang-etal-2022-large,bib-data-leakage-2023} and {\modified{generally in generative neural networks \cite{bib-unintended-memorization-2019}.}} 
{\modified{In response to privacy and security concerns surrounding these models, some authors initially developed methods to decrease the total quantity of memorized text \cite{bib-remove-memorization1, bib-remove-memorization2,Carlini2023}. These approaches were later reconsidered because, alongside "bad" memorization, there is also "good" memorization. The latter involves accurately recalling factual events and details, helping to avoid generating plausible but incorrect information or "hallucinations" \cite{bib:predictable-memorization-2023}, such as the case of our paper.}} 
{\modified{In the same direction, a very recent work \cite{bib-pretraining-detection2-2024} presented a method to estimate the degree of dataset contamination and copyrighted books in LLMs, based on the hypothesis that an example unseen during training given in input to the detection method tends to contain a few outlier words with low probabilities, whereas a seen example is less likely to contain words with such low probabilities. \citet{bib-pretraining-detection-2024} investigates the same problem by examining next-word-prediction-based language models with membership adversarial attacks to determine whether a given text was included in pre-training data. \citet{bib-black-box-llm-2024} addressed the same issue using a reference model and a classifier. To detect whether a text was used during pre-training, they first memorize it with a reference model. The prefix of the text is then fed into a LLM to generate a continuation. Both the original text and the generated text are input into the reference model to extract language modeling probabilities. If the text was part of the LLM's pre-training, the generated continuation closely matches the original, leading to high probabilities. A classifier then uses these features to determine if the text was used in pre-training. 
}}

{\modified{All the above papers mainly focus on \textit{verbatim} detection in LLMs. 
In contrast, \citet{bib-entity-level-memorization-2024} explored the concept of entity memorization, noting that verbatim detection methods might miss certain memorized content when key or sensitive information is embedded only in specific parts of the data. Specifically, they define entity memorization as occurring when a model is prompted with input derived from partial entities within the training data, and the model's output correctly includes the expected entity information. In this work, we complement their study of memorization using a broad definition of entity with a more classic, ontology-centric view of entities as concepts from the vocabulary of a reference ontology.}}

Recently, \citet{Ishihara2023} conducts a comprehensive survey on training data extraction from Pre-trained Language Models (PLMs), providing a taxonomy of memorization definitions and systemizing approaches for both attack and defense. In our work, we follow this line of research while focusing on \emph{memorization of information} rather than textual tokens, entities, and sequences.
%
%
He et al.\ \cite{Horrocks2023} proposed to investigate PLMs' knowledge of ontologies using a set of inference-based probing tasks, thus showing that these models encode little subsumption relations. 
{\modified{\citet{bib-knowleadgeable-llms-kg-2024} also demonstrated that LLMs still struggle with accurately grasping factual knowledge, as revealed by their analysis of LLMs' memorization of certain knowledge-graph relationships. For this reason, the \citet{triplet} proposed to combine triplets-based searches in a semi-structured knowledge base with the generative capacities of LLMs to improve their answers. Finally, the \citet{bib:-PLMs-understand-onto2023} tested BERT- and RoBERTa-based language models' ability to memorize and reason with ontological knowledge, finding that they struggle on these tasks.
A common challenge observed in these tasks is the tendency of LLMs to generate hallucinations, i.e., statements that are plausible and contextually coherent but factually incorrect \cite{hallucination0}. This issue has been widely studied, both proposing detection methods and evaluation metrics with a rapidly growing body of literature \cite[\emph{inter alia}]{hallucination1,hallucination2,hallucination3}. Our work also contributes to this field by assessing LLMs' hallucination tendencies when interacting with ontologies.
}}
In this work, we focus primarily on the terminology component and its provenance with respect to the training data. The importance of our work stems from the fact that PLMs are becoming ever more critical for ontology-centric tasks like ontology mapping and alignment \citep{Paulheim2023, heiko-2024}, whose evaluation may also suffer from dataset leakage \citep{sainz-etal-2023-nlp,deng2023investigating}.

\section{Do LLMs dream of ontologies?}
\label{sec:task}

\subsection{Task description}\label{sub:task-description}

To evaluate the memorization of ontological information in LLMs, we propose a simple task inspired by that of \textit{exact memorization} \citep{exact-memorization}.  
In particular, we ask the LLMs to return in a zero-shot fashion the ID of an ontological concept, given in input, through a natural language prompt, the concept's label. 
No further training on domain data is thus performed, and no information is inserted into the LLMs context other than that in the prompt.

We remark that the ID-concept label association is very basic information contained in (many) ontologies, which typically include more complex and structured knowledge -- e.g., concept taxonomies, concept relations, and axioms \citep{Horrocks2023}: however, IDs and concept labels are (i) inherently sequences of characters and (ii) frequently mentioned together in textual content (e.g., Webpages, scientific articles), thus making the ID-concept label association something likely to be observed in the training material of LLMs and thus helpful in evaluating memorization in LLMs.

\subsection{Resources}  \label{sub:ontologies}

\paragraph{Ontologies.}
We opt for ontologies whose entities\footnote{In this work, use the terms \textit{entity} and \textit{concept} interchangeably.} are uniquely identified with an ID syntactically unrelated to the entity's label to follow a similar approach to that of \textit{exact memorization} analysis. Doing so guarantees that if the LLM associates the correct ID to an input label, it is because the ID and the corresponding label have been encountered in the training material and thus have been memorized, as it is impossible to derive the ID from the label only. 
{\modified{We focus on three domain ontologies (Gene Ontology (GO) \citep{geneontology}, Uberon Ontology \citep{uberon}, and ICD-10 \cite{ICD-10}) because domain-specific terms tend to have a lower polysemy \citep{koeling-etal-2005-domain} and thus allow us to ignore issues related to label disambiguation (i.e., cases where the same term can be associated with multiple IDs). 
Moreover, these ontologies and their associated concepts are extensively referenced in scientific literature, much of which is accessible on the Web and can be utilized as training data for LLMs.
We, however, also test the performance on a general-purpose ontology-based resource, i.e., Wikidata \cite{wikidata}, to study the memorization of LLMs in a more complex setting, including polysemy to provide a complete range of experiments.}}

GO provides a computational representation of current scientific knowledge about the functions of protein and non-coding RNA molecules produced by genes from many different organisms. It is widely used to support scientific research and is thus popular in Web publications.
It contains $42,854$ different concepts, each having a label and uniquely identified by a numerical ID prefixed by "\texttt{GO:}" (a.k.a., GO ID). For example, the label \textit{kidney development} is associated with the concept whose GO ID is \texttt{GO:0001822}, while the label \textit{kidney morphogenesis} to GO ID \texttt{GO:0060993}. 
As evident from the examples, without memorization, it would be impossible for the model to derive the correct GO ID given only the label. 
Uberon is an integrated cross-species anatomy ontology representing a variety of entities classified according to anatomical structure, function, and developmental lineage. In the Uberon ontology, there are $15,543$ terms (a.k.a., labels) that are uniquely associated with a numerical UBERON ID that is specific to the ontology, 
i.e., identified by the prefix \texttt{"UBERON"}. The ontology also incorporates terms from other ontologies (e.g., GO and BFO), which we have excluded from the analysis as they are not native to Uberon. While the Uberon ontology is more recent and less prevalent in the scientific literature than GO,\footnote{To quantify, on December 19, 2023, the GO's paper (\cite{geneontology}) is cited on Google Scholar $38,553$ times, while the Uberon's paper (\cite{uberon}) $686$ times.} it allows us to compare performance across different resources and subdomains, i.e., proteins vs. anatomy.
{\modified{The third analyzed resource is ICD-10,  a very popular and globally recognized classification system\footnote{Classification systems, such as thesauri, are generally regarded as \emph{lightweight} ontologies~\cite{Giunchiglia2009}.} developed by the World Health Organization (WHO) for coding and classifying diseases, injuries, and various health conditions. As for Gene and Uberon, each concept is associated with a unique alphanumeric code. It is widely used by healthcare providers, insurers, and health organizations to maintain records, track diseases, and collect data for health statistics. While GO and Uberon belong to the biomedical domain, ICD-10 belongs to the clinical one.
Of all the concepts present in ICD-10 ($73,201$), we consider only those whose ID has no more than four digits, that is, those that describe diseases that are not too specific, obtaining a list of $11,494$ concepts. This choice is mainly made to reduce the costs of the experiments.}}
{\modified{The fourth analyzed resource is Wikidata, a general-purpose knowledge base that stores concepts by unique identifiers. Analyzing all the concepts in Wikidata (approximately $100M$) would be very expensive and time-consuming. For this reason, a representative subsample of $30K$ concepts is extracted, having different popularity according to QRank, a measure that aggregates page view statistics for estimating the popularity of a concept.\footnote{Wikidata QRank: \url{https://github.com/brawer/wikidata-qrank} [Last access on October 7, 2024]} The Wikidata setting is more complex because, being a general-purpose resource, it may have cases of polysemy (different IDs having the same label) that have to be addressed during the evaluation phase.
}}

\paragraph{Large Language Models.} \label{sub:llm}
We consider {\modified{four}} different LLMs. The first one is the 12 billion parameters release of EleutherAI's Pythia \citep{Pythia2023} (named \pythia now on). It is trained on ThePile dataset \citep{thePile}, comprising 825 GiB of English text from different sources, including PubMed Central (90.27 GiB of data) and PubMed Abstract (19.26 GiB), which are biomedical portals and thus can include literature inherent to the Gene and Uberon ontologies. {\modified{Although initially ThePile was publicly released, it was later withdrawn, and at the time of these experiments, it is now no longer accessible, thus preventing the possibility of more in-depth analyses}}. {\modified{The second analyzed LLM is Google GEMINI-1.5-Flash (named \gemini now on), an API-accessible LLM whose training data is not publicly available nor declared}}.
Finally, we test two OpenAI chatbot models,\footnote{\url{https://openai.com/chatgpt} [Last access on October 7, 2024]} i.e., GPT-3.5-Turbo-0613 (named \gptt now on) and GPT-4-0613 (called \gptf now on) respectively. 
The details of both \gptt and \gptf training datasets are not publicly available.

\subsection{Research Questions} \label{sub:researchq}
To study the degree of memorization of ontology IDs and labels in LLMs, we address the following research questions:
\begin{description}
    \item[RQ1] Can LLMs correctly predict the ID of the concepts in a known ontology, only given the concepts' label in the input prompt? Are there differences in performance between the LLMs analyzed? Are there any common patterns in errors made by LLMs? {\modified{What is the impact of hallucination on models' performance? This RQ is addressed in Section \ref{sec:rq1}.}}
    \item[RQ2] Does the accuracy of the ID prediction correlate with the number of times the ID-concept label association is found on the Web and thus (likely) present and frequent in the training material? Does the latter also influence errors made by LLMs? {\modified{This RQ is addressed in Section \ref{sec:rq2}.}}
    \item[RQ3] 
    How does the observed consistency of ID prediction for a given prompted label, under various prompt repetitions or perturbations, inform our understanding of the memorization of the ID-concept label association?
    {\modified{This RQ is addressed in Section \ref{sec:rq3}.}}
\end{description}
The first two research questions investigate whether, how, and why LLMs have memorized some basic ontological information. The third one looks more closely into ways to empirically assess the extent of memorization in the model of such ontological information by observing whether the prediction of the LLMs changes when prompting it multiple times for the same information.

\section{Memorization of ontologies (RQ1)} \label{sec:rq1}

\subsection{Methodology}
\label{sub:methodrq1}

To address RQ1, we evaluate the performance of the considered LLMs on the prediction task described in Section~\ref{sub:task-description} for all concepts defined in each ontology presented in Section~\ref{sub:ontologies}. We calculate the accuracy as the proportion of all instances where the model returns the exact correct ID for a given concept’s label over all the instances in the ontology (e.g., for the label \textit{kidney development} from GO, the prediction is considered correct only if the ID is precisely \texttt{GO:0001822}). 
For Wikidata, where polysemy can occur, we consider a predicted ID correct if it belongs to the set of IDs whose label matches the one specified in the prompt.

\paragraph{Quantifying error patterns.} To analyze possible common patterns in errors made by LLMs on the given prediction task, i.e., predicting a sequence of digits (the concept ID) given one or more word tokens (the concept label), we propose to investigate whether some \textit{syntactic} similarity (e.g., character or token-based) exists between the gold and wrongly-predicted IDs, or the corresponding concept labels, following two complementary strategies.

The first approach investigates if the model tends to make mistakes by providing an ID similar to the correct one. For example, the GO ID \texttt{GO:0060219} (\textit{camera-type eye photoreceptor cell differentiation}) is close to \texttt{GO:0060519} (\textit{cell adhesion involved in prostatic bud elongation}) because only one digit needs to be changed to go from the first to the second. 
To do this analysis, we extract the list of wrong IDs predicted by the models and the gold one and compute the Levenshtein Distance \citep{Levenshtein} on them.

The second approach investigates if the model tends to make mistakes by providing an ID whose corresponding label is similar to the correct one. For example, the label \textit{heart valve morphogenesis} (GO ID \texttt{GO:0003179}) is close to \textit{heart trabecula morphogenesis} (GO ID \texttt{GO:0061384}) due to two words in common, or the label \textit{regulation of cell division} (GO ID \texttt{GO:0051302}) with \textit{regulation of cell motility } (GO ID \texttt{GO:2000145}) because of three words in common. In these examples, the numerical IDs of the two concepts are very different, but the label has sub-tokens in common, which could lead to a prediction error by the model.
To do this analysis,  we extract the list of errors committed by the models by collecting the gold label (i.e., the correct one) and the label corresponding to the wrong ID predicted by the model. Then, we divide the two strings into tokens and compute the Jaccard Similarity \citep{Jaccard} on them.

\paragraph{Model hallucinations.} 
We also check if the models tend to predict IDs that do exist in the ontology or if they sometimes make up the proposed IDs.
To do so, we first collect all the unique IDs predicted by the model and determine how many are invented, i.e, not present in the corresponding ontology.\footnote{In the cases where the prediction task described in Section~\ref{sub:task-description} is assessed on a subset of the original ontology (i.e., ICD-10 and Wikidata), an ID is considered invented only if it is not present in the entire original ontology (rather than just the subset).} {\modifiedblue{Additionally, we calculate the percentage of incorrect predictions where the model invents a plausible, yet non-existent, ID}}.

\subsection{Experimental setup} \label{sub:experimentsrq1}
For RQ1, we assessed the accuracy of \pythia, \gemini, \gptt, and \gptf on GO, Uberon, ICD-10, and Wikidata.
For each model, we first tried slightly different prompts on a sample of 100 random entities, achieving comparable performance across the variations.\footnote{All considered prompts are available in the GitHub repository of the work.} Therefore, we selected the ones with the least number of input tokens to reduce experimentation costs for paid models. For chatbots models (\gptt, \gptf and \gemini), the following prompts were selected:
\begin{itemize}
    \item \textit{Provide the GO ID for the label "{l}". In the answer write only the corresponding GO ID.}
    \item \textit{Provide the UBERON ID for the label "{l}". In the answer write only the corresponding UBERON ID.}
    \item \textit{Provide the ICD-10 ID for the label "{l}". In the answer write only the corresponding ICD-10 ID.}
    \item \textit{Provide the Wikidata ID for the label "{l}". In the answer write only the corresponding Wikidata ID.}
\end{itemize}
while for \pythia, we selected the following text completion prompts (requesting the addition of maximum $10$ new tokens): 
\begin{itemize}
    \item \textit{In the Gene Ontology, the GO ID of the label "{l}" is GO:}
    \item \textit{In the Uberon Ontology, the Uberon ID of the label   "{l}" is UBERON:}
    \item \textit{In the ICD-10, the ICD-10 ID of the label  "{l}" is}
    \item \textit{In the Wikidata, the Wikidata ID of the label  "{l}" is}
\end{itemize}

All the prompts are executed with the temperature set to $0.0$.\footnote{We used a regular expression to extract only the concept ID from responses to ensure an accurate evaluation of the outputs. This approach allows us to handle outputs containing additional text, such as "The ID is GO:XXXX" by isolating the relevant ID and avoiding unnecessary penalties. As a result, such responses are not discarded or penalized, thereby mitigating potential issues related to output constraint violations, which have been highlighted as a concern for recent OpenAI models \cite{rdf}.}

\subsection{Results}\label{sub:resultrq1}
Table~\ref{tab:performance} reports the accuracy of the four compared models on the task described in Section \ref{sub:task-description}, for the four considered ontologies. 
Table~\ref{tab:errors} summarizes the resulting Jaccard Similarity and Levenshtein distance, averaged over all the wrong predictions by the models on all considered ontologies. {\modifiedblue{Finally, Table~\ref{tab:uniqueids} provides details about model hallucinations: Table~\ref{tab:unique_invented} summarizes the number of unique IDs produced by the models across the different ontology settings, along with the percentage of IDs invented by the model (i.e., non-existent IDs in the ontology), while Table~\ref{tab:percentage_error} shows the percentage of wrong predictions caused by invented IDs.}}

\begin{table}[tb] 
\caption{Accuracy of \pythia, \gemini, \gptt and \gptf on the task described in Section \ref{sub:task-description}, on the Gene Ontology, Uberon, ICD-10 and Wikidata datasets. The score of the best-performing model on each ontology is in bold.}
\label{tab:performance}
\addtolength{\tabcolsep}{-0.0pt} 
{\begin{tabular}{p{4.0cm}|rrrr}
\hline
Dataset & \pythia & \gemini & \gptt & \gptf\\
\hline
Gene Ontology& .0067 & .0140 & .0611 & \textbf{.1270}  \\
Uberon Ontology& .0000 & .0003 & .0035  & \textbf{.0129}   
\\
ICD-10 & .0061 & .0648 & .2487  & \textbf{.3749} 
\\
Wikidata & .0000 & .0001 & .0026  & \textbf{.0073}  \\
\hline
\end{tabular}}
\end{table}

\begin{table}[tb] 
\caption{Average similarity measures of wrongly predicted IDs (Levenshtein distance, abbreviated with Levenshtein d.) and associated concept labels (Jaccard similarity, abbreviated with Jaccard s.) for \pythia, \gemini, \gptt and \gptf models, on Gene Ontology (GO), Uberon Ontology (UO), ICD-10 (ICD) and Wikidata (WD).}
\label{tab:errors}
\addtolength{\tabcolsep}{-0.0pt} 
{\begin{tabular}{p{4.0cm}|rrrr}
\hline
Measure & \pythia & \gemini & \gptt & \gptf\\
\hline
Levenshtein d. \footnotesize{(GO)} & 4.409 
& 4.317 & 3.964  & 
3.814  \\
Jaccard s. \footnotesize{(GO)} & .149 & .186 & .301 
& .338  \\
\hline
Levenshtein d. \footnotesize{(UO)} & 6.570 & 4.048 & 3.533  & 4.580   \\
Jaccard s. \footnotesize{(UO)} & .001 & .016 & .047  & .033  \\
\hline
Levenshtein d. \footnotesize{(ICD)} & 5.348  & 5.339  & 5.387 & 5.382  \\
Jaccard s. \footnotesize{(ICD)} & 
$\approx .000$
& .001 & .004  & .008  \\
\hline
Levenshtein d. \footnotesize{(WD)} & 4.593  & 5.039  & 4.669 & 4.662  \\
Jaccard s. \footnotesize{(WD)} & 
$\approx .000$
& .003 & .001  & .014  \\
\hline
\end{tabular}}
\end{table}


\begin{table}[tb] 
    \caption{(\ref{tab:unique_invented}): number of unique IDs (Unq) produced by the models (\pythia, \gemini, \gptt, and \gptf) for Gene Ontology (GO), Uberon Ontology (UO), ICD-10 (ICD), and Wikidata (WD), and percentage of unique IDs being invented, {\modifiedblue{i.e., hallucinated}}, by the model (\%Inv); (\ref{tab:percentage_error}): percentage of wrong predictions caused by invented IDs. Next to each ontology in the subtable \ref{tab:unique_invented}, the number of expected unique IDs is reported.}
    \label{tab:uniqueids}
    \begin{minipage}{\textwidth}
    \centering
        \begin{minipage}{\textwidth}
            \centering
            \subcaptionbox{Unique predicted IDs and percentage of them being invented {\modifiedblue{due to hallucination}}.\label{tab:unique_invented}}{
                \begin{tabular}{r|rrr|rrr|rrr|rrr}
                    \hline
                    & \multicolumn{3}{c|}{\pythia} & \multicolumn{3}{c|}{\gemini} & \multicolumn{3}{c|}{\gptt} & \multicolumn{3}{c}{\gptf} \\
                    & Unq & \%Inv & & Unq & \%Inv & & Unq & \%Inv & & Unq & \%Inv & \\
                    \hline
                    GO (42,854) & 1,353 & 10.64 & & 5,558 & 16.61 & & 7,182 & 7.64 & & 12,308 & 9.60 \\
                    UO (15,543) & 30   & 30.00 & & 237  & 5.06  & & 641  & 10.92 & & 2,971  & 33.52 \\
                    ICD (11,494) & 1,002 & 61.98 & & 3,288 & 44.65 & & 5,593 & 27.50 & & 6,633  & 20.08 \\
                    WD (30,000) & 125  & 26.40 & & 8,817 & 28.16 & & 7,734 & 16.89 & & 15,539 & 19.31 \\
                    \hline
                \end{tabular}
            }
        \end{minipage}
        \vspace{0.5em}
        \\ 
        \begin{minipage}{\textwidth}
        \centering
            \subcaptionbox{Percentage of wrong predictions caused by invented IDs {\modifiedblue{due to hallucination}}.\label{tab:percentage_error}}{
                \begin{tabular}{r|r|r|r|r}
                    \hline
                    & \pythia & \gemini & \gptt & \gptf \\
                    \hline
                    GO \phantom{00,000} & 6.85 & 11.84  & 6.35 & 7.63  \\
                    UO \phantom{00,000}  & 12.10 &  1.96  &  2.34 &  15.94  \\
                    ICD \phantom{00,000}  & 53.62 &  43.51 & 32.59 & 30.18  \\
                    WD \phantom{00,000}  & 2.76  &  28.91 & 11.55 &  15.13  \\
                    \hline
                \end{tabular}
            }
        \end{minipage}
    \end{minipage}
\end{table}

\subsection{Discussion} \label{sub:discussionsrq1}

Table \ref{tab:performance} shows that
\gptf and \gptt are partially familiar with both ICD-10 and GO by obtaining an accuracy of $.37$ and $.25$ respectively on the first and $.13$ and $.06$ respectively on the second. \gemini and \pythia reach on ICD-10 and GO lower performance than \gptt and \gptf, but they still know ICD-10 and GO more than Wikidata and Uberon. 
On all the datasets, \gptf has an accuracy higher than all the other models: this result aligns with the findings of other papers on LLMs tokens memorization, e.g., \citet{Carlini2023}. \pythia achieves the lower accuracy scores in all datasets 
and gets close to zero accuracy in both Wikidata and Uberon. \gemini also achieves an accuracy that is nearly zero on them, but its performance on ICD-10 and GO is much higher than \pythia's.
%
We can thus observe that:
\begin{itemize}
    \item {\modifiedblue{Overall}}, the accuracy scores are quite low for all models and ontologies considered, with the exception of \gptt and \gptf on ICD-10;
    \item {\modifiedblue{For}} all datasets, \gptf and \gptt perform better than the others, with \gptf being the best-performing LLM, followed by \gptt and \gemini, with \pythia achieving the lowest scores: a possible explanation is the different number of model parameters (much higher for \gptf and \gptt than \gemini and \pythia) or the different amount of data used in the training phase:\footnote{We recall that, while for \pythia the dataset used for training is declared (but not available anymore), for \gptt, \gptf, and \gemini models it is not.} for instance, the zero accuracy of \pythia on Uberon can be ascribed to the almost total absence of Uberon IDs in its training dataset ($15,494$ of the $15,543$ Uberon IDs are not present at all in PubMed, a website used to train \pythia), which instead contains occurrences of several Gene Ontology and ICD-10 IDs;
    \item {\modifiedblue{The}} performances on the Uberon Ontology and Wikidata are lower than on the ICD-10 and Gene Ontology for all the considered models: a possible explanation could be that in ICD-10 and GO, the IDs are more frequently used in practice by domain experts than Uberon and Wikidata ones, and thus the concept ID-label pairs for the latter may have been seen far less in the training material of the models (more on this in Section~\ref{sec:rq2}).
\end{itemize}

Table \ref{tab:errors} summarizes the results of the error analysis. Concerning the Levenshtein distance of the wrongly predicted IDs with respect to the gold ones, the difference in performance across the model is rather minimal on all considered ontologies. This may be because all models tend to predict IDs that, even if wrong, have a format comparable with the gold ones, especially in terms of used characters and length. The differences across the ontologies are also quite minimal, ranging from $3.814$ on the Gene Ontology (with \gptf) to $6.570$ on Uberon (with \pythia). Looking at the Jaccard Similarity scores, the models achieving the highest values when making wrong predictions are \gptf and \gptt, followed by \gemini and \pythia, meaning that the former tend to predict (wrong) IDs whose associated concept labels are more similar to the one corresponding to gold IDs than the latter. Overall, the Jaccard Similarity of all models is much higher in Gene Ontology than the other ontologies. 

Concerning models' hallucinations, Table \ref{tab:uniqueids} first shows that all models tend to produce only a portion of the expected IDs for the considered ontologies, thus indicating that each model repeatedly predicts the same (wrong) IDs for different concept labels (Table \ref{tab:unique_invented}). The lower numbers of unique IDs across all ontology settings are generated by \pythia, meaning that this model tends to predict few, frequently repeated IDs for multiple labels, while \gptf is by far the model producing the higher number of different IDs for each ontology. The models literally invent several of the wrongly predicted IDs. The percentage values (\%Inv) of invented IDs over the produced ones actually vary a lot across models and ontologies, ranging from $5.06\%$ for \gemini on Uberon to $61.98\%$ for \pythia on ICD-10. For \pythia, \gemini, and \gptt, the highest proportion of invented IDs occur on ICD-10 (resp., $61.98\%$, $44.65\%$, and $27.50\%$), while for \gptf on Uberon ($33.52\%$). Except for \gemini, the models invent proportionally fewer unique IDs on the Gene Ontology. Finally, looking at the percentage of errors due to invented IDs (Table \ref{tab:percentage_error}), the highest values are observed on ICD-10 for all models, ranging from $30.18\%$ for \gptf to $53.62\%$ for \pythia. 
{\modifiedblue{The results indicate that hallucinations significantly impact model performance, likely due to issues in the training data, the training process, or inference. While beyond the scope of our work, we acknowledge that mitigation methods (e.g., \cite{bib:hallucination-new}) could be applied to partially address the model’s hallucinations.}}

\vspace{1em}
\noindent
\textbf{Answer to RQ1:} the results show that the considered LLMs have rather minimal knowledge of the information in the considered ontologies. \gptf substantially outperforms all the other analyzed models in terms of accuracy, and even when making wrong predictions, the errors made are generally closer to the gold one than the other models. Finally, all models exhibit some form of hallucination when performing the task.

\section{Memorization and popularity on the Web (RQ2)} \label{sec:rq2}

\subsection{Methodology}
\label{sub:methodrq2}

RQ2 aims to assess the correlation (if any) between the popularity of a concept on the Web\,---\,and by extension, its likely presence in LLMs’ training material\,---\,and the model’s memorization of it. It is important to note that precise information on the actual training data used for certain models (namely \gptt, \gptf, and \gemini) is not publicly available. Additionally, the dataset used to train \pythia, known as ThePile, is no longer accessible. However, much of ThePile consists of publicly available Web material, including sources like PubMed, and it is well-established that models such as \gptf, \gptt, and \gemini were trained, at least in part, on large volumes of Web content \citep{Bamman2023}.

To explore this relationship, we construct a dataset that includes information on the popularity of ontology $(ID, label)$ pairs on the Web. Popularity is approximated by the number of times the pair appears together in the same Web documents. To estimate this, we use Google Search APIs with the query format ``label''  ``ID'' to obtain exact matches from Web content. While the implementation details of Google Search are not publicly disclosed, making it a black-box tool \citep{kilgarriff-2007-last}, it is widely used for similar research purposes \citep{Bamman2023}.

For the analysis, we group the ontology concepts into buckets based on the popularity of their $(ID, label)$ pairs on the Web. We then examine how the model’s average accuracy in predicting IDs varies across these buckets (i.e., the proportion of correctly predicted IDs for the concepts within each bucket). To assess potential correlations between Web popularity and model accuracy, we apply Spearman’s rank correlation~\cite{spearman04}. Additionally, we evaluate whether Web popularity Granger-causes~\cite{bib:granger} model accuracy, i.e., whether a series of values for the former helps predict the latter.
\footnote{While typically applied to time series data, the concept of Granger causality can also be applied to cross-sectional data \cite{Lu_Su_White_2017}, i.e., collected at a single point in time, provided that there is a clear ordering or progression of the data, in our cases given by the ranking of Web occurrences. Similar considerations also apply to later uses of Granger causality in the paper.}

\paragraph{Web occurrences vs.\ PubMed occurrences} To provide some further evidence for our hypothesis that Web occurrences are a viable way to approximate the distribution of ontology $(ID, label)$ pairs in the training material of LLMs, we assess the correlation and Granger causality between the occurrences of domain-specific ontology concepts (Gene Ontology) in the Web and in a domain-specific resource (PubMed) that was used, among other non-domain-specific resources (and thus unlikely to contain substantial information about the given ontology), to train one of the considered LLMs (\pythia). We use the Google Search APIs, restricted to the PubMed domain, to collect the required dataset-specific $(ID, label)$ occurrences for the considered ontology. Spearman's rank statistical correlation and Granger causality are computed between the distribution of Web occurrences and PubMed occurrences, as well as between PubMed occurrences and the \pythia's prediction accuracy. The statistical significance of the values is computed according to the Permutation test.

\paragraph{Error patterns vs.\ popularity.} We also complement the study of possible common patterns in mistakes made by LLMs on the given prediction task (cf.\ Section~\ref{sub:methodrq1}) with further analysis related to the popularity of the concepts. 
First, we check whether the similarity between wrong and gold predictions based on Levenshtein distance and Jaccard similarity correlates with the number of Web occurrences of the concepts. Then, we investigate if the model is biased towards popular concepts, i.e., if it tends to answer with the ID of a very popular concept. For the latter, we extract the most common incorrectly predicted IDs and check which buckets they belong to estimate if there is a Spearman's rank statistical correlation 
between frequently predicted IDs and their bucket number, with statistical significance computed according to the Permutation test.

Moreover, we further investigate the repeated IDs phenomenon discussed at the end of Section~\ref{sub:resultrq1} in light of the popularity data. We collect the top-$k$ most repeated IDs for each model for the Gene Ontology and check if they are uniformly distributed in the buckets according to the actual distribution of the concepts in the buckets. More in detail, for each bucket $B_{i}$, we compute:
\[
R_{B_{i}} = \frac{n_{B_{i}}}{\frac{N_{B_{i}}}{N} \cdot k}
\]
where $n_{B_{i}}$ is the number of top-$k$ repeated IDs that belong to bucket ${B_{i}}$, $N_{B_{i}}$ is the total number of IDs in bucket ${B_{i}}$, and $N$ is the total number of IDs across all buckets.

\subsection{Experimental setup} \label{sub:experimentsrq2}
We computed the Web occurrences for all datasets considered in our work (Gene Ontology, Uberon, ICD-10, and Wikidata). However, given the very low performance on the Uberon and Wikidata ontologies for all models -- cf.\ Section~\ref{sub:resultrq1}, we considered only Gene Ontology and ICD-10 to address RQ2.
To group elements into buckets based on Web popularity, we calculated the 50 percentiles of the observed Web occurrences and allocated the ontology elements into the corresponding frequency-based buckets. Given the distribution of occurrences in Gene Ontology and ICD-10, the first buckets, corresponding to less frequent $(ID, label)$ pairs, contain much more elements than the last ones, where very frequent $(ID, label)$ are allocated. We arbitrarily set the value to $50$ to ensure that the obtained buckets have a substantial number of elements to conduct the analysis. Similarly, $50$ buckets were also created to organize the PubMed occurrences of the Gene Ontology, to study the correlation between Web occurrences and PubMed occurrences, and between PubMed occurrences and the \pythia's prediction accuracy. To study the relation between repeated IDs and popularity we empirically set $k=500$, thus considering the top-500 most repeated IDs for each model. All the prompts are executed with the temperature set to $0.0$.

\subsection{Results}\label{sub:resultrq2}
\begin{figure}[!t]
    \centering
    \includegraphics[width=0.9\columnwidth]{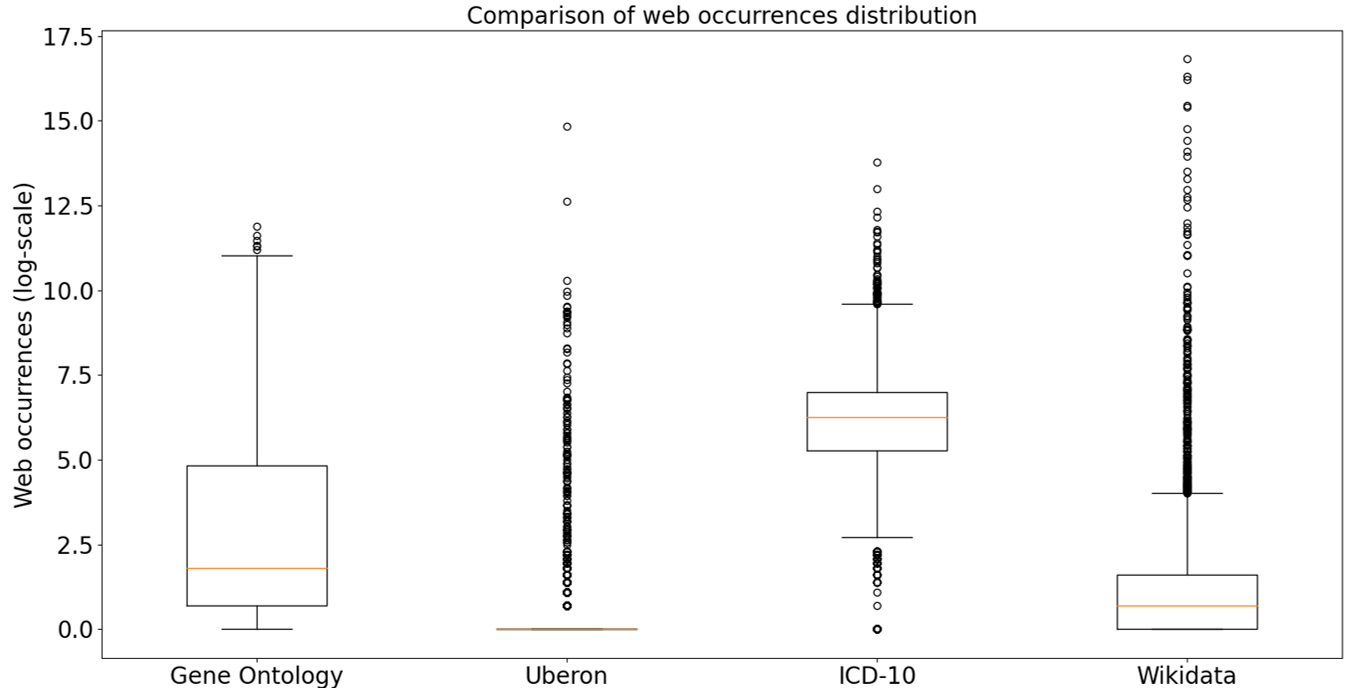}
    \caption{Distribution of the number of Web occurrences (log-scaled, base $e$) for each dataset (Gene Ontology, Uberon, ICD-10, and Wikidata).}
    \label{fig:occurrences_distribution}
\end{figure}

\begin{figure}[!t]
    \centering
    \begin{subfigure}{0.49\textwidth}
        \centering
        \includegraphics[width=1.0\textwidth]{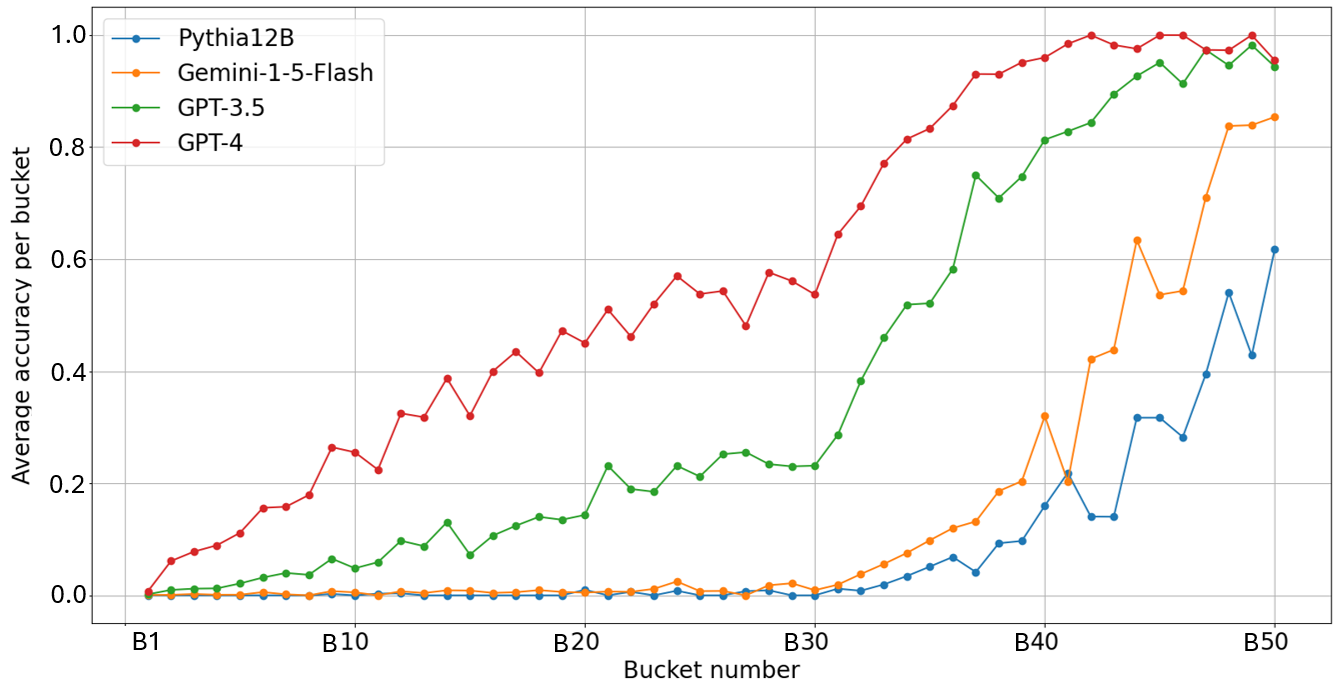}
        \caption{\footnotesize{Gene Ontology}}
        \label{fig:go_dataset_occurrences_combined}
    \end{subfigure}
    \begin{subfigure}{0.49\textwidth}    
        \centering
        \includegraphics[width=1.0\textwidth]{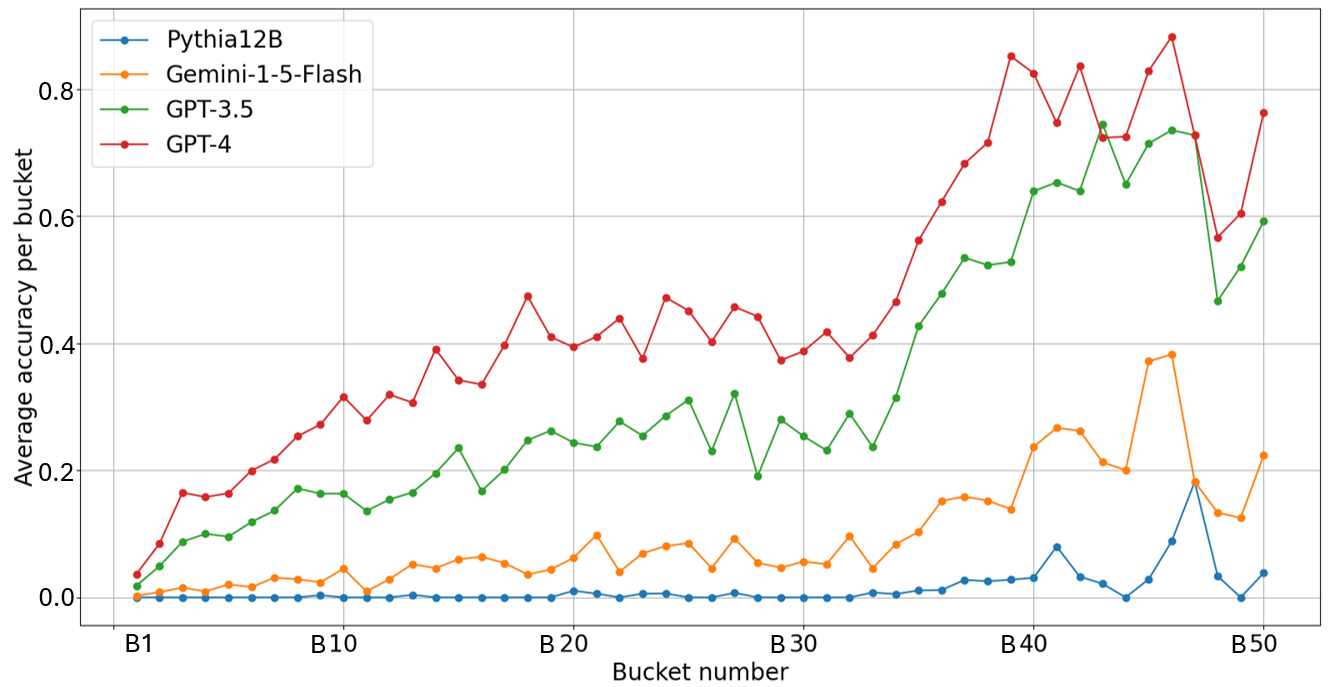}
        \caption{\footnotesize{ICD-10}}
        \label{fig:ICD_dataset_occurrences_combined_bucket_number}
    \end{subfigure}
    \caption{
        Average accuracy of the model's ID prediction (y-axis) according to the popularity of the concept on the Web (represented by the bucket number in x-axis) in the Gene Ontology (a) and ICD-10 (b).}
        \label{fig:dataset_occurrences_combined_main}
\end{figure}

Figure ~\ref{fig:occurrences_distribution} reports the popularity of ID-label pairs for all considered datasets on the Web.  
The two plots in Figure~\ref{fig:dataset_occurrences_combined_main} 
show how the accuracy of the ID predictions of the models, based on the prompted concept label for the Gene Ontology (Figure~\ref{fig:go_dataset_occurrences_combined}) and ICD-10 (Figure~\ref{fig:ICD_dataset_occurrences_combined_bucket_number}), varies according to how frequently that $(ID, label)$ pair occurs on the Web. 
IDs were organized in buckets, labeled from B1 (least frequent $(ID, label)$ pairs - rarely observed) to B50 (most frequent $(ID, label)$ pairs - observed tens of thousands of times) based on the number of occurrences in the Web. 
Table~\ref{tab:acc_correlation_causality} reports the resulting Spearman's rank correlation coefficient (Table \ref{tab:correlation}) and the Granger causality F-statistic (Table \ref{tab:causality}) between the number of Web occurrences of the $(ID, label)$ pairs and the accuracy of all the models for the Gene Ontology and ICD-10. 


\begin{table}[tb] 
    \caption{Spearman's rank correlation coefficient (\ref{tab:correlation}) and Granger causality F-statistic with lag=3 (\ref{tab:causality}) between the number of Web occurrences of the $(ID, label)$ pairs and the accuracy of the models for the Gene Ontology and ICD-10. Scores marked with ``*'' means the reported value is statistically significant (p-value $\leq .05$).}
    \label{tab:acc_correlation_causality}
    
    \begin{minipage}{\textwidth}
        \centering
        
        \subcaptionbox{Correlation Results\label{tab:correlation}}{
            \begin{tabular}{p{4.0cm}|rrrr}
                \hline
                & Pythia-12B & GEMINI-1.5 & \gptt & \gptf\\
                \hline
                GO Correlation  & $.850$* & $.924$* & $.993$*  & $.982$*   \\
                ICD-10 Correlation  & $.692$* & $.901$* & $.933$*  & $.919$*   \\
                \hline
            \end{tabular}
        }
        \vspace{0.5em} 
        \subcaptionbox{Causality Results\label{tab:causality}}{
            \begin{tabular}{p{4.0cm}|rrrr}
                \hline
                & Pythia-12B & GEMINI-1.5 & \gptt & \gptf\\
                \hline
                GO Causality  & $6.753*$ & $7.498$* & $3.336*$ & $2.853$*  \\
                ICD-10 Causality  & $30.338*$ & $1.089\phantom{*}$ & $1.711\phantom{*}$ & $2.592 \phantom{*}$  \\
                \hline
            \end{tabular}
        }
    \end{minipage}
\end{table}

\paragraph{Web occurrences vs.\ PubMed occurrences} Table~\ref{tab:pudmed_cor_caus} reports the resulting Spearman's rank correlation coefficient and the Granger causality F-statistic between (i) the number of Web occurrences and the number of PubMed occurrences and (ii) the number of PubMed occurrences of the $(ID, label)$ pairs and the accuracy of \pythia's prediction for the Gene Ontology.
\begin{table}[tb] 
\caption{Spearman's rank correlation coefficient and the Granger causality F-statistic (lag=3)  between the number of Web occurrences and the number of PubMed occurrences (first column) and the number of PubMed occurrences of the $(ID, label)$ pairs and the accuracy of \pythia's prediction for the Gene Ontology (second column). Scores marked with ``*'' means the reported value is statistically significant (p-value $\leq .05$).}
\label{tab:pudmed_cor_caus}
\addtolength{\tabcolsep}{-0.5pt} 
{\begin{tabular}{p{4.0cm}|rrrr}
\hline
 & Web occurrences vs. & PubMed occurrences vs. \\
 & PubMed occurrences & \pythia accuracy \\
\hline
GO Correlation  & .796* & .865* \\
GO Causality  & 527.386* & 6.825* \\
\hline
\end{tabular}
}
\end{table}

\begin{figure}[!t]
    \centering
    \begin{subfigure}{0.49\textwidth}
        \centering
        \includegraphics[width=1.0\textwidth]{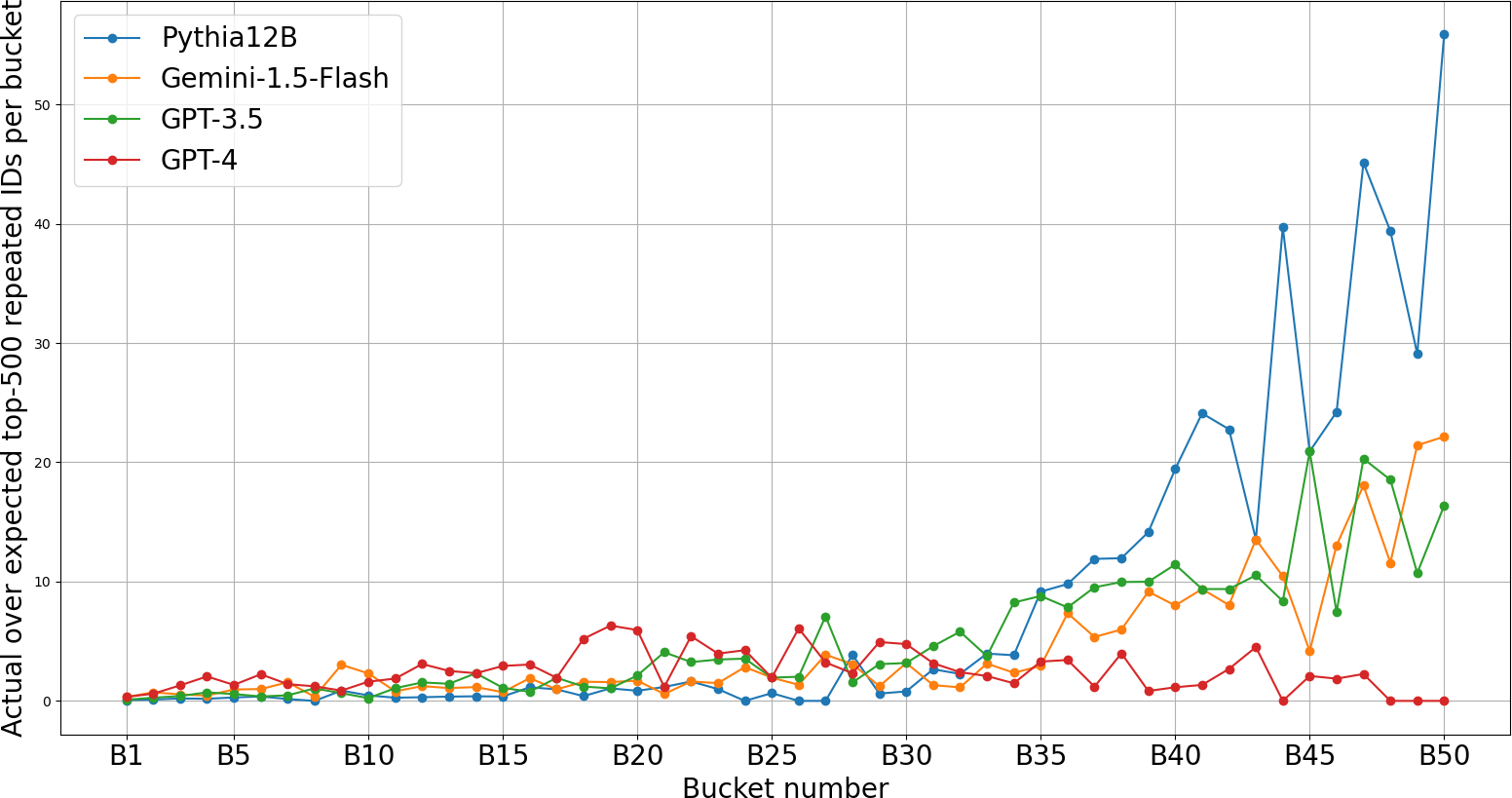}
        \caption{\footnotesize{Gene Ontology}}
        \label{fig:bias-go}
    \end{subfigure}
    \hfill
    \begin{subfigure}{0.49\textwidth}    
        \centering
        \includegraphics[width=1.0\textwidth]{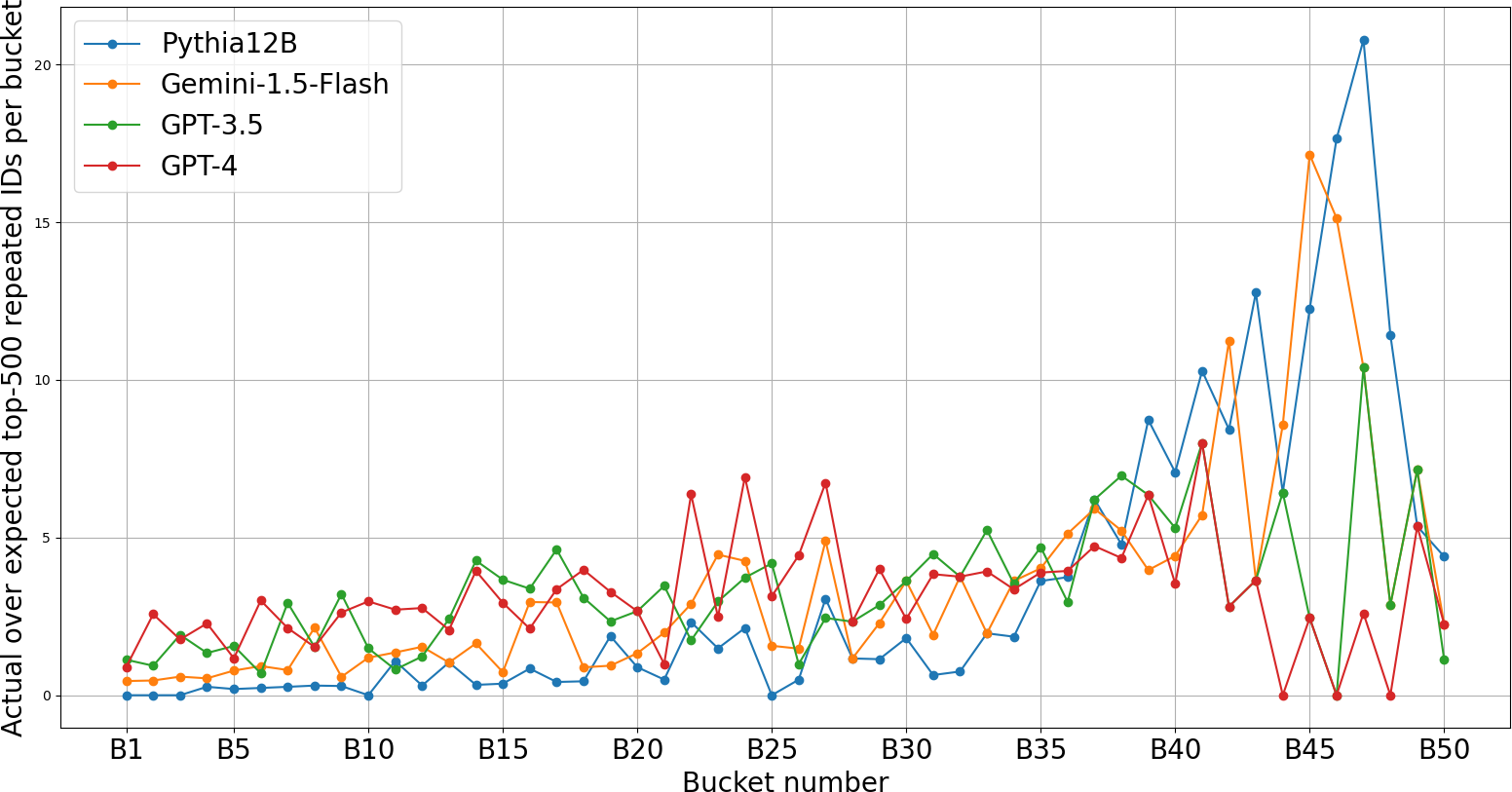}
        \caption{\footnotesize{ICD-10}}
        \label{fig:bias-icd}
    \end{subfigure}
    \caption{Variation over all the buckets of the $R_{B_{i}}$ values, capturing the ratio between the number of top-500 repeated IDs that belong to bucket $B_{i}$, and the proportion of them that should be in the bucket $B_{i}$ according to the overall distribution of all IDs in the buckets for Gene Ontology (\ref{fig:bias-go}) and ICD-10 (\ref{fig:bias-icd}).}   
        \label{fig:bias}
\end{figure}

\paragraph{Error patterns vs.\ popularity.} Table \ref{tab:error_correlation} reports the Spearman's rank correlation coefficient between the ranking of the buckets according to the averaged Levenshtein distance / Jaccard similarity per bucket and the ranking of the buckets.
\begin{table}[tb] 
\caption{Spearman's rank correlation coefficient between the Levenshtein distance / Jaccard similarity measures and the ranking of the buckets according to the averaged number of Web occurrences of the $(ID, label)$ pairs in the bucket for Gene Ontology (GO) and ICD-10. Only wrong predictions are considered. Scores marked with ``*'' means the correlation is statistically significant (p-value $\leq .05$).}
\label{tab:error_correlation}
\addtolength{\tabcolsep}{-0.5pt} 
{\begin{tabular}{p{4.0cm}|rrrr}
\hline
Similarity Measure & \pythia & GEMINI-1.5 & \gptt & \gptf\\
\hline
(GO) Levenshtein d.  & -.714* & -.853* & -.364*  & -.208\phantom{*}   \\
(GO) Jaccard s.  & .788* & .874* & .590* & .595*  \\
\hline
(ICD-10) Levenshtein d.  & -.948* & -.957* & -.945*  & -.926*   \\
(ICD-10) Jaccard s.  & .082\phantom{*} & .233\phantom{*}& .242\phantom{*}& .310*  \\
\hline
\end{tabular}
}
\end{table}
Finally, the plots in Figure~\ref{fig:bias} show the ratio $R_{B_{i}}$ between actual and expected top-500 repeated IDs for all buckets $B_{i}$ for both Gene Ontology (Figure \ref{fig:bias-go}) and ICD-10 (Figure \ref{fig:bias-icd}).
%
%
For Gene Ontology, a strong correlation was observed for \pythia (.868), \gemini (.871), and \gptt (.947), both statistically significant, while no correlation was detected for \gptf. For ICD-10, a strong correlation was observed for \pythia (.901) and \gemini (.827), both statistically significant, while no correlation was detected for \gptt and \gptf. 

\subsection{Discussion} \label{sub:discussionsrq2}
Figure~\ref{fig:occurrences_distribution} shows that ICD-10 is the dataset with the highest median number of documents on the Web containing both the ID and label for its concepts, followed by Gene Ontology, Wikidata, and Uberon. Partly surprisingly given its wide adoption and its broad general-domain coverage, Wikidata IDs and labels are infrequently used together in the same Web documents.
In contrast, IDs and labels of domain-specific resources like ICD-10 and Gene Ontology are more frequently used together. This opposite situation may be because ICD-10 and Gene Ontology are highly specialized ontologies used extensively in their respective fields (clinical medicine and biology), where precise and standardized terminology is crucial for accurate communication and documentation, while Wikidata entities may be referred to in many cases only by their label (or description) without explicit reference to the specific ID.

Comparing these values with the accuracy score of the various LLMs on the ontologies reported in Table~\ref{tab:performance}, this may indeed suggest that the more some concept $(ID, label)$ pairs are seen in Web documents (and, thus, likely in training material of LLMs), the more the information is memorized by the models, and thus correctly predicted when prompted about it. Indeed, this connection between Web occurrences and accuracy finds more confirmation in Figure \ref{fig:dataset_occurrences_combined_main}, which shows how the accuracy score of all the models vary based on the number of Web occurrences of the GO and ICD-10 concepts. Indeed, with minimal variations, we observe that accuracy substantially improves as we move from low-frequency concepts to high-frequency ones. That is, the plots show that the considered models tend to make correct predictions for concepts frequently observed on the Web (and, thus, likely commonly observed in the training material), and wrong predictions for less frequently observed concepts, thus suggesting a certain degree of correlation between the accuracy of the predictions (averaged per bucket), i.e., the degree of concepts correct memorization, and the occurrence of the concepts in the Web. Indeed, for GO, this is confirmed by the correlation and Granger causality scores reported in Table~\ref{tab:acc_correlation_causality}: there is a strong correlation (statistically confirmed for all LLMs) between Web occurrences and the model's accuracy, with the former that Granger causes the latter (statistically confirmed for all LLMs). For ICD-10, there is a strong and statistically relevant correlation between Web occurrences and the model's accuracy (statistically confirmed for all LLMs), with the former that Granger causes the latter for \pythia.

The accuracy score of \gptf on (almost) every bucket is higher than that of \gptt, which in turn is higher than that of \gemini, which is higher than that of \pythia. For some high-frequency buckets, \gptf scores reach perfect accuracy on the Gene Ontology, and above .800 on ICD-10.

\paragraph{Web occurrences vs.\ PubMed occurrences} The results in Table~\ref{tab:pudmed_cor_caus} confirm a strong correlation between Web and PubMed occurrences for the Gene Ontology $(ID, label)$ pairs 
with the latter that Granger causes the former,
thus suggesting that, despite differences in the absolute value of the occurrences (the average number of occurrences for a GO concept on the Web is approximately 50 times higher than in PubMed), their overall distributions are comparable. Indeed, Table~\ref{tab:pudmed_cor_caus} also shows that a strong correlation exists between PubMed occurrences and \pythia's prediction accuracy, with the former that Granger causes the latter, exactly as observed for Web occurrences (cf. Table~\ref{tab:acc_correlation_causality}).
Even taking into account all the limitations previously mentioned, we believe this is yet another indication that Web occurrences could be taken as a reasonable approximation of the training material of LLMs in all those (many) cases where no information on the latter is provided, at least for the considered task.

\paragraph{Error patterns vs.\ popularity.} The results in Table~\ref{tab:error_correlation} show that in most of the considered model/onto\-logy configurations, when the models make prediction errors, the wrongly predicted IDs are syntactically closer to the gold one (directly or indirectly through the associated labels) for concepts frequently occurring on the Web rather than infrequent ones. Indeed, a strong inverse correlation is observed between Levenshtein distance and Web occurrences for both ontologies and for all models, except for \gptt (moderate) and \gptf (poor) on Gene Ontology, all statistically significant (except for \gptf). The situation is more varying for the Jaccard similarity, where moderate to strong correlation (statistically significant) is observed for all models on the Gene Ontology, while the correlation is poor/moderate for all models on ICD-10. Again, these findings further suggest that the more the $(ID, label)$ association occurs on the Web, the more the models are likely closer to making the correct prediction. 

Concerning the experiments on the distribution of the top-500 repeated IDs, the plot for the Gene Ontology (Figure~\ref{fig:bias-go}) clearly shows that for \pythia, \gemini, and \gptt, most of them proportionally belong to the buckets containing the most occurring $(ID, label)$ pairs on the Web, while for \gptf these IDs seem to be more proportionally distributed across all buckets. Indeed, the reported correlation values confirm that \pythia, \gemini, and \gptt are somehow biased toward frequently occurring concepts on the Web, while this does not hold for \gptf. Although moderately noisier, especially toward the top-most high-frequency buckets (B48-B50), the plot for ICD-10 (Figure \ref{fig:bias-icd}), shows a similar trend as for the Gene Ontology. In this case, the reported correlation values confirm that \pythia and \gemini are more biased toward frequently occurring concepts on the Web, while this does not hold for \gptt and \gptf. That is, when making errors, the less-performing (\pythia and \gemini) models seem to repeatedly predict the IDs-label that they have likely seen the most in the training data, while the better-performing models are marginally (\gptt) or not affected at all (\gptf) by this.

\vspace{1em}
\noindent
\textbf{Answer to RQ2:} the results show that the prediction accuracy of the considered LLMs substantially correlates with the occurrence of the $(ID, label)$ associations in Web documents: the more the latter, the better the former. Indeed, under the assumption that these models are trained on a vast amount of Web content, and thus that Web occurrences can be taken as a good approximation of occurrences in the training material, these findings suggest that the more some information is seen in the training material, the more it is memorized by the models, as shown by the prediction accuracy as well as the way the models fail in predicting the correct IDs.

\section{Assessing the relationship between prompt invariance and memorization (RQ3)} \label{sec:rq3}

\subsection{Methodology}
\label{sub:promptinvariance}

To assess to which extent the $(ID, label)$ association for an ontological concept is correctly memorized in an LLM (cf. RQ3), we propose using the model’s response invariance as a metric. We suggest measuring the variability in the model’s answers when the prompt is repeatedly submitted or perturbed in different ways to indicate how well the correct ID for a given label is memorized. Specifically, we investigate three different strategies applied to the base prompts selected in Section~\ref{sub:experimentsrq1}:
\begin{itemize}
    \item \textbf{PI-1}: we repeat the same prompt more times {\modifiedblue{asking for determinism in the answer and counting the number of different answers.}} Intuitively, if the $(ID, label)$ association is well memorized in the model, only one answer should always be returned.
    \item \textbf{PI-2}: we repeat the same prompt {\modifiedblue{adjusting the degree of randomness in the model’s output, ranging from deterministic to more creative behavior.}} Intuitively, if the model well memorizes the $(ID, label)$ association, the answer should remain consistent regardless of the temperature, meaning it should not change even as the model’s creativity increases.
    \item \textbf{PI-3}: we repeat the same prompt in different languages {\modifiedblue{asking for determinism.}} In this implementation, we translate only the prompt while we keep the ontology's labels in the original language, i.e., English. Intuitively, if the $(ID, label)$ association is well memorized in the model, the answer should be independent of the language used to make the request.
\end{itemize}

In detail, given a set of $N$ buckets $B_{i}$, with $i \in [1,2,...., N]$, each containing $K$ random concepts $C_{Bi} = [c_{1}, c_{2}, ..., c_{K}]$,\footnote{While the approach could be applied to all concepts of an ontology, we limited their number to economize on the costs of the LLM APIs.} we evaluate the performance by considering the mean number $U$ of unique answers returned by the model for concepts belonging to each bucket $B_{i}$. In particular, for each concept $c_{j} \in B_{i}$, we compute the \textit{Prediction Invariance} after $M$ prompt repetitions as:
\[
PI_{c_{j}}^{M} = 1 - \left(\frac{U-1}{M-1}\right),
\]
{\modifiedblue{where $U$ is the number of unique answers returned after $M$ prompts. 
If the $M$ prompts return always the same answer for a given $c_{j}$, $PI_{c_{j}}$ will be $1.0$ (i.e., max prediction invariance). If the $M$ prompts return $M$ different answers for a given $c_{j}$, $PI_{c_{j}}$ will be $0.0$ (i.e., minimum prediction invariance).}} We then average the different $PI_{c_{j}}$ obtained for each concepts $c_{j} \in B_{i}$, obtaining the \textit{Average Prediction Invariance} for each bucket $B_{i}$ as: 
\[
AvPI_{(B_{i})} = \frac{1}{K} \sum_{c_{j}\in C_{B_{i}}} PI_{c_j}^{M}.
\]
Finally, we use Spearman's rank correlation coefficient to study the relationship between $AvPI_{(B_{i})}$ and accuracy for the same bucket $B_{i}$ (i.e., the correctness of the predictions).

\subsection{Experimental setup}\label{sub:expertimentalrq3}

Similar to RQ2, we address this research question using the Gene Ontology and ICD-10 datasets. For the LLMs, we conduct the tests with \gptt, \gemini, and \pythia.\footnote{To minimize experiment costs, we excluded \gptf; however, initial assessments indicated a similar trend to that observed with \gptt.} Starting from the same $N=50$ buckets used for the previous experiments, we considered $K=20$ random IDs per bucket, for a total of $1,000$ instances, allowing us to study the impact of the three invariance strategies on the model's behavior according to the popularity of the concepts. In PI-1, the same prompt is repeated $M=10$ times with the temperature set to zero. In PI-2, we tested $M=11$ different temperature levels, from $0.0$ to $1.0$ with $0.1$ increments. In PI-3, we tested $M=5$ different languages, namely English, Italian, German, French, and Spanish. The prompts were manually translated by people proficient in the respective languages.

\subsection{Results}\label{sub:resultrq3}

\begin{figure}[!t]
    \centering
    \begin{subfigure}{0.49\textwidth}
        \centering
        \includegraphics[width=1.0\textwidth]{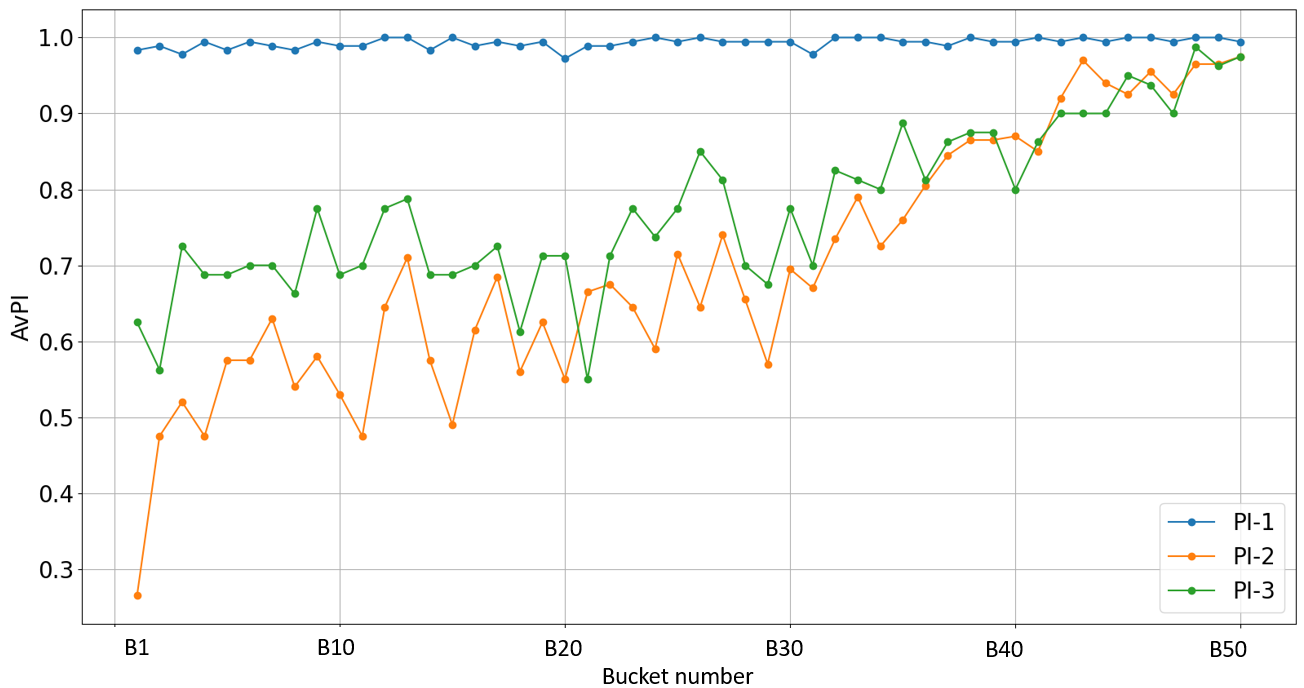}
        \caption{AvPI per bucket - \gptt for GO}
        \label{fig:PI-GO-GP3}
    \end{subfigure}
    \hfill
    \begin{subfigure}{0.49\textwidth}    
        \centering
        \includegraphics[width=1.0\textwidth]{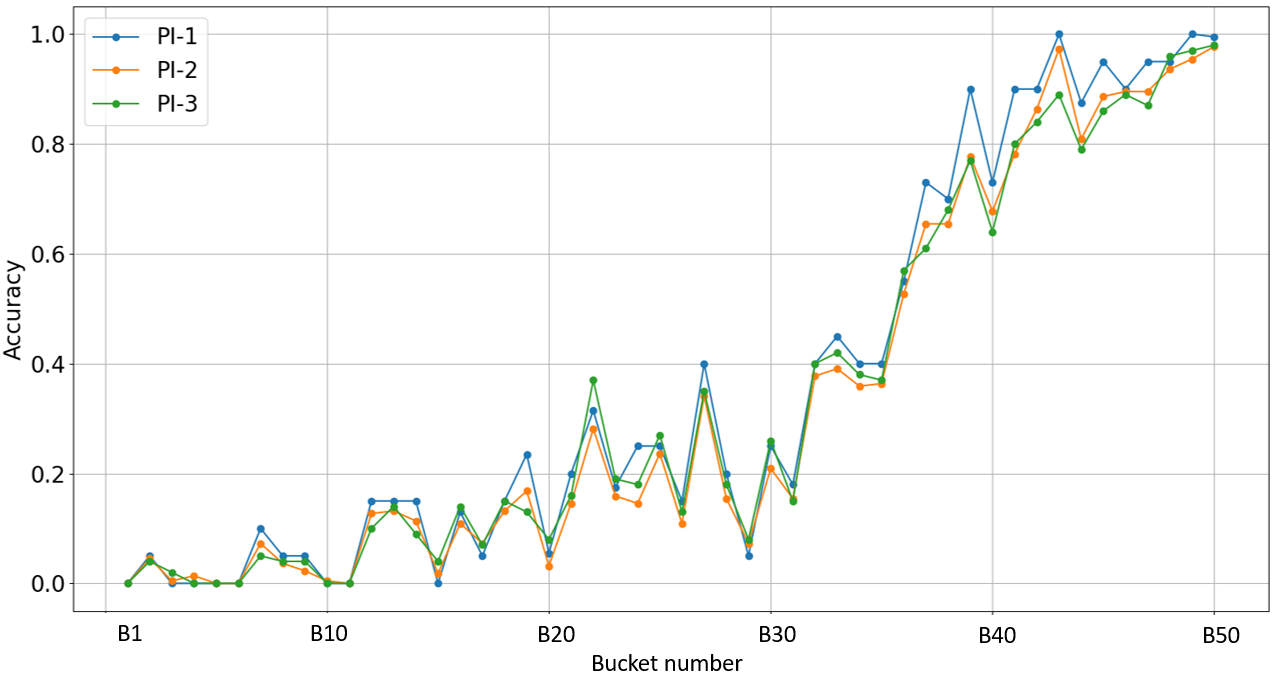}
        \caption{Accuracy ($\%$) per bucket - \gptt for GO}
        \label{fig:PI2-GO-GP3}
    \end{subfigure}
    \begin{subfigure}{0.49\textwidth}
        \centering
        \includegraphics[width=1.0\textwidth]{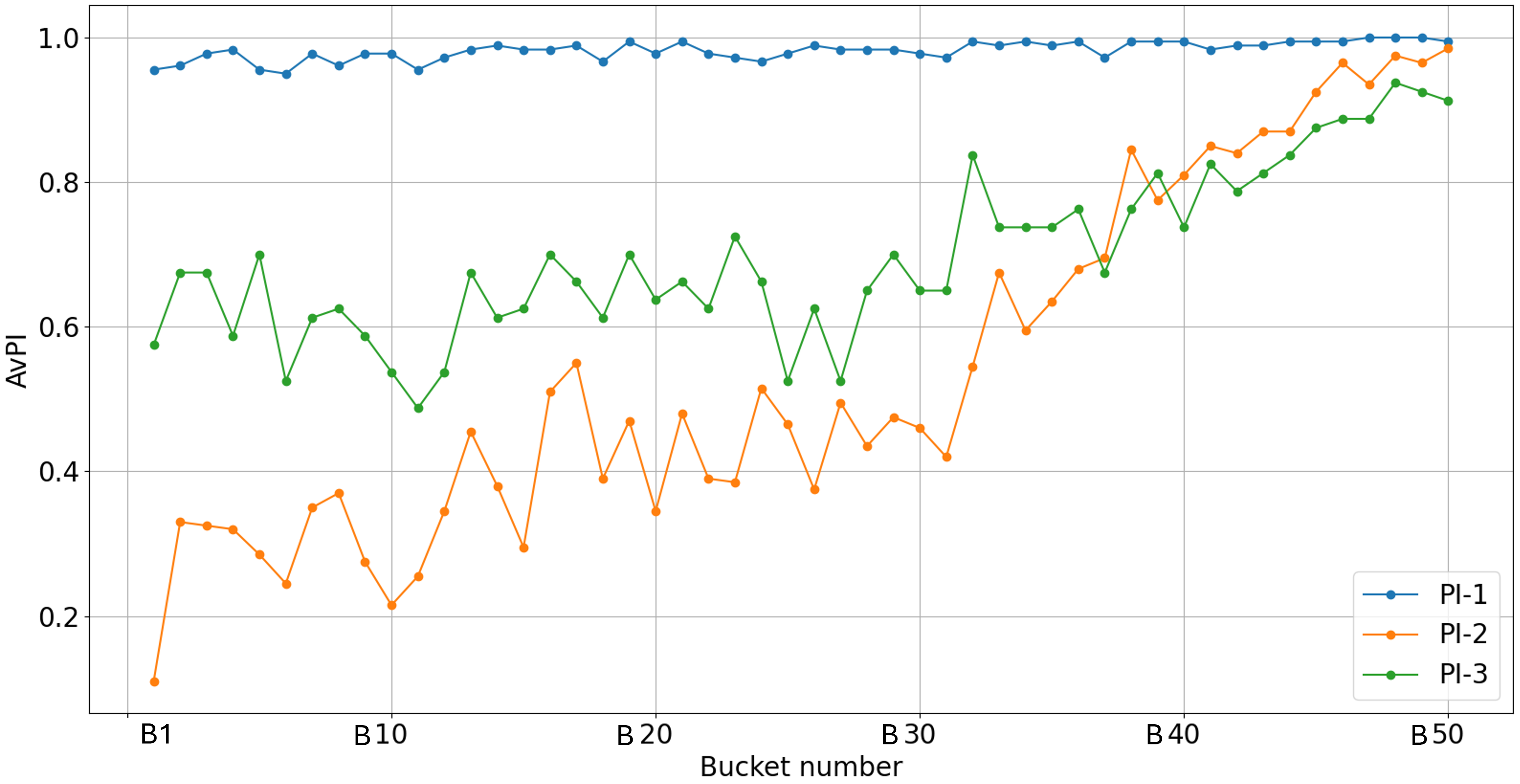}
        \caption{AvPI per bucket - GEMINI for GO}
        \label{fig:PI-GO-GEM}
    \end{subfigure}
    \hfill
    \begin{subfigure}{0.49\textwidth}    
        \centering
        \includegraphics[width=1.0\textwidth]{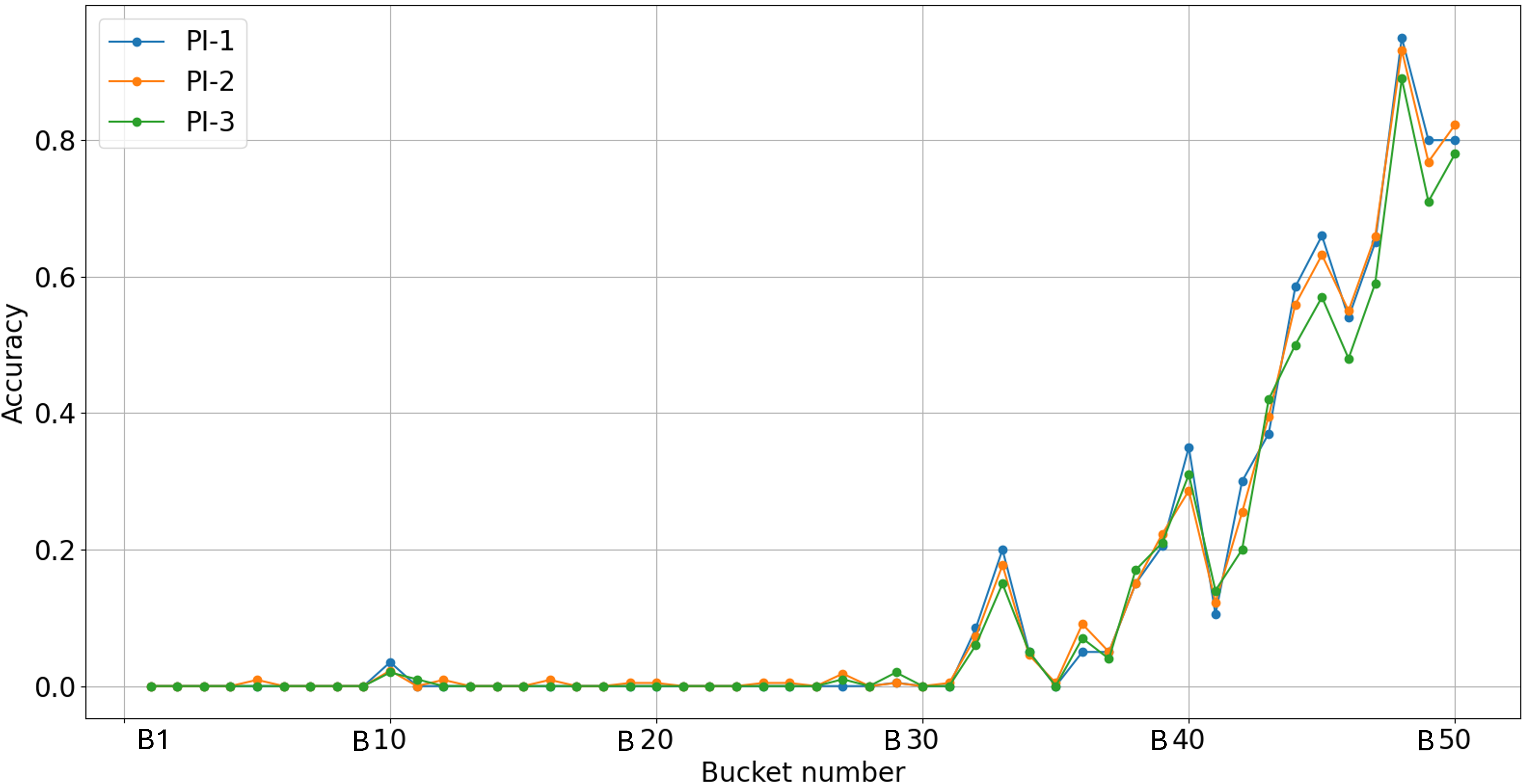}
        \caption{Accuracy ($\%$) per bucket - GEMINI for GO}
        \label{fig:PI2-GO-GEM}
    \end{subfigure}
    \begin{subfigure}{0.49\textwidth}
        \centering
        \includegraphics[width=1.0\textwidth]{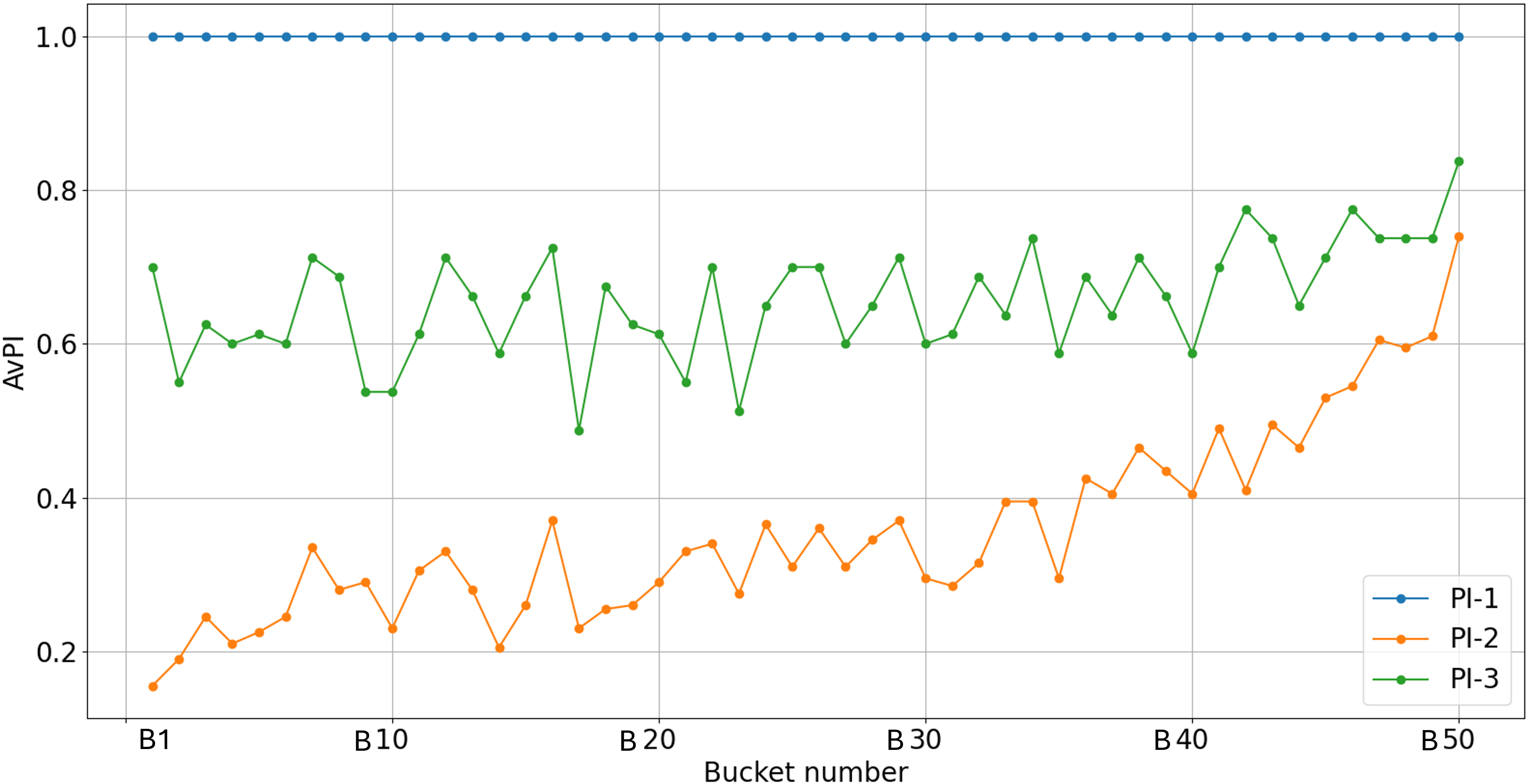}
        \caption{AvPI per bucket - Pythia for GO}
        \label{fig:PI-GO-PYT}
    \end{subfigure}
    \hfill
    \begin{subfigure}{0.49\textwidth}    
        \centering
        \includegraphics[width=1.0\textwidth]{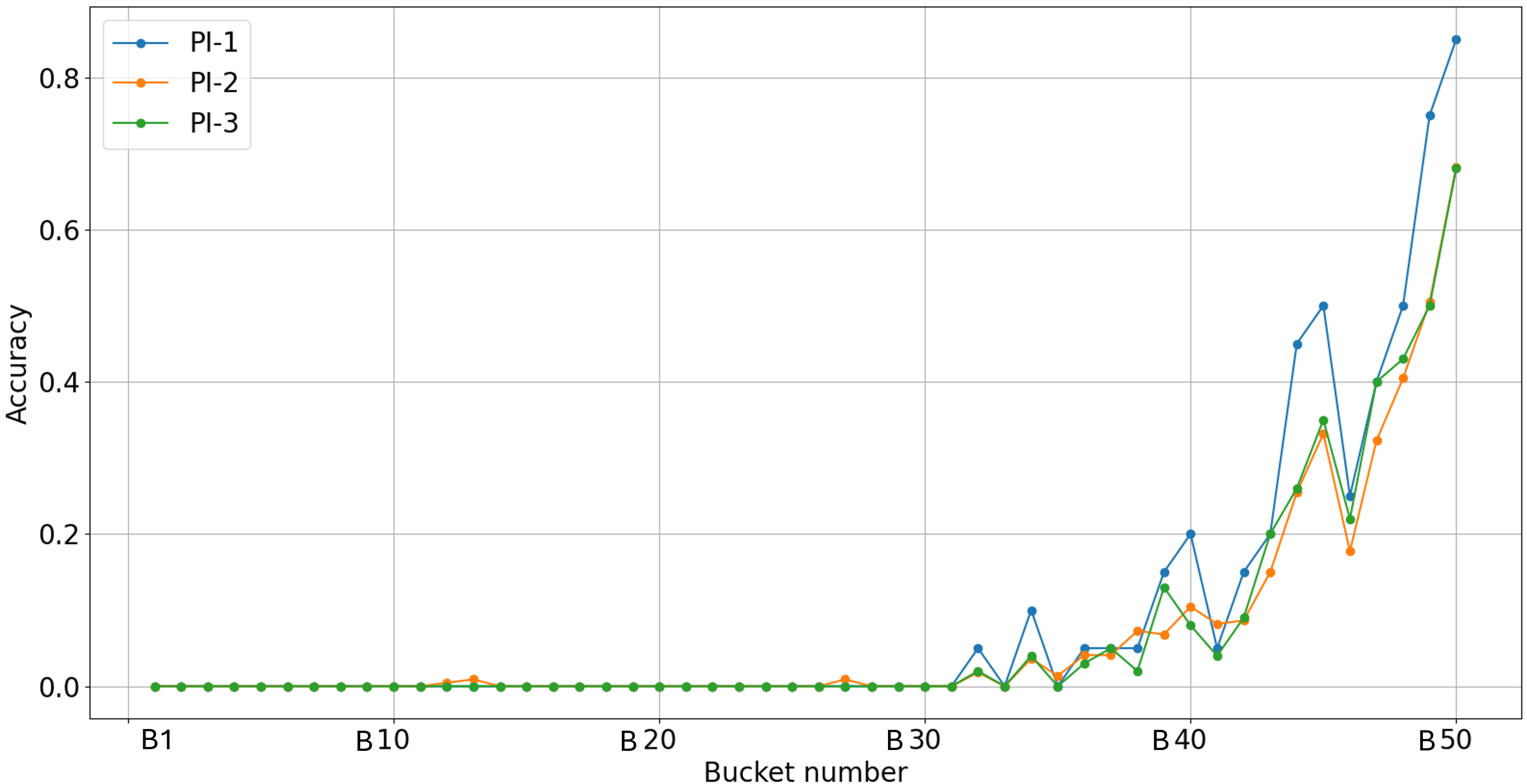}
        \caption{Accuracy ($\%$) per bucket - Pythia for GO}
        \label{fig:PI2-GO-PYT}
    \end{subfigure}
    \caption{
        Variation of AvPI (left) and accuracy (right) on the different buckets of the Gene Ontology when applying the PI-1, PI-2, and PI-3 invariance strategies to \gptt (Subfigures \ref{fig:PI-GO-GP3}-\ref{fig:PI2-GO-GP3}), \gemini (Subfigures \ref{fig:PI-GO-GEM}-\ref{fig:PI2-GO-GEM}) and \pythia (Subfigures \ref{fig:PI-GO-PYT}-\ref{fig:PI2-GO-PYT})
        }    
    \label{fig:RQ3-per-bucket}
\end{figure}

Figure~\ref{fig:RQ3-per-bucket} shows the trend of $AvPI$ (left) and accuracy (right) for \gptt, \gemini, and \pythia on the Gene Ontology. The Spearman's rank coefficient between AvPI and accuracy is high (moderate to very strong correlation) for both PI-2 and PI-3 for \gptt ($.950$ and $.850$, respectively), \gemini ($.813$ and $.743$, respectively), and \pythia ($.795$ and $.581$, respectively), all statistically confirmed ($p<.05$) by the Permutation test. No statistical correlation is observed for PI-1.


\begin{figure}[!t]
    \centering
    \begin{subfigure}{0.49\textwidth}
        \centering
        \includegraphics[width=1.0\textwidth]{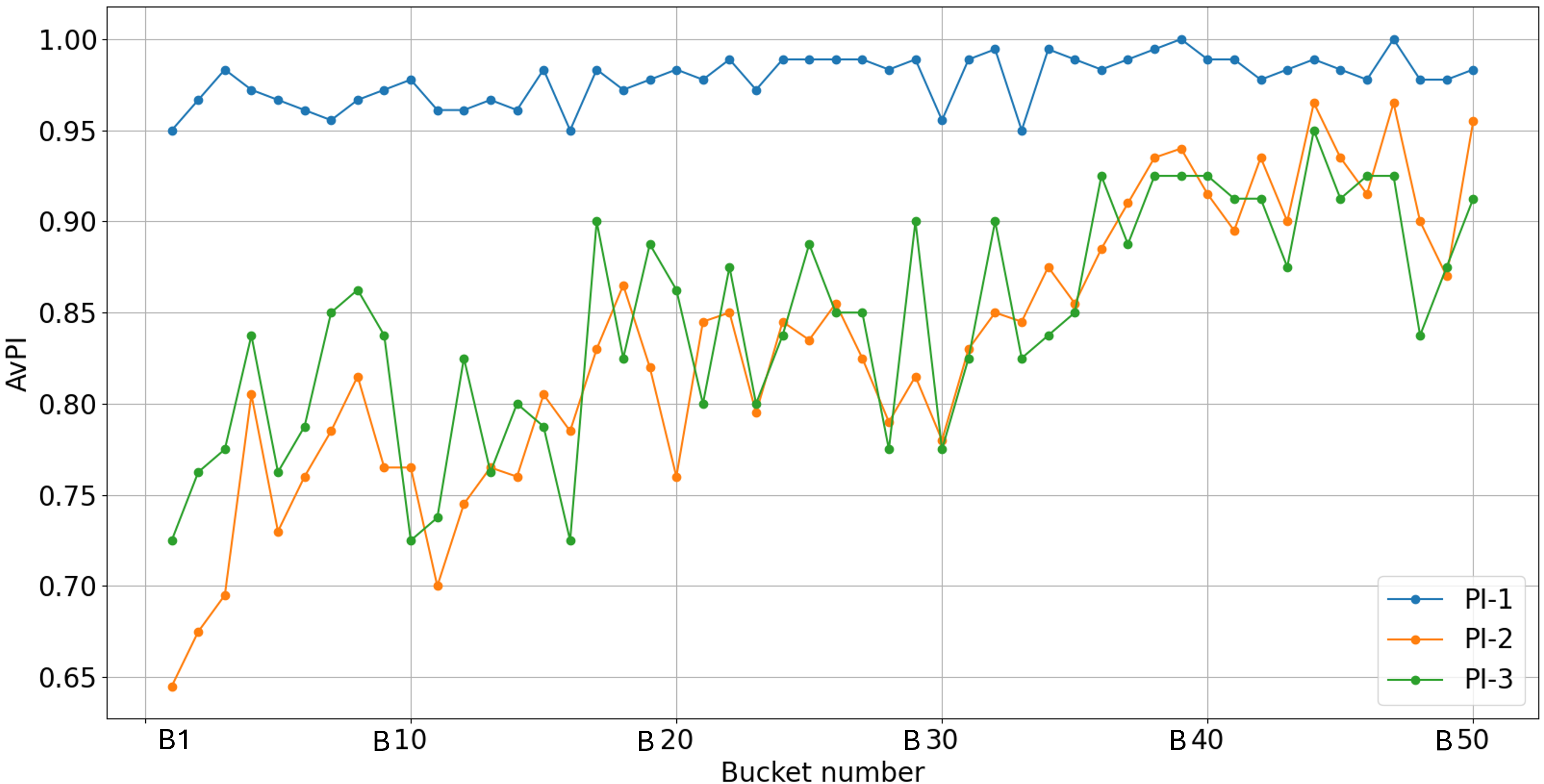}
        \caption{AvPI per bucket - \gptt for ICD}
        \label{fig:PI-ICD-GP3}
    \end{subfigure}
    \hfill
    \begin{subfigure}{0.49\textwidth}    
        \centering
        \includegraphics[width=1.0\textwidth]{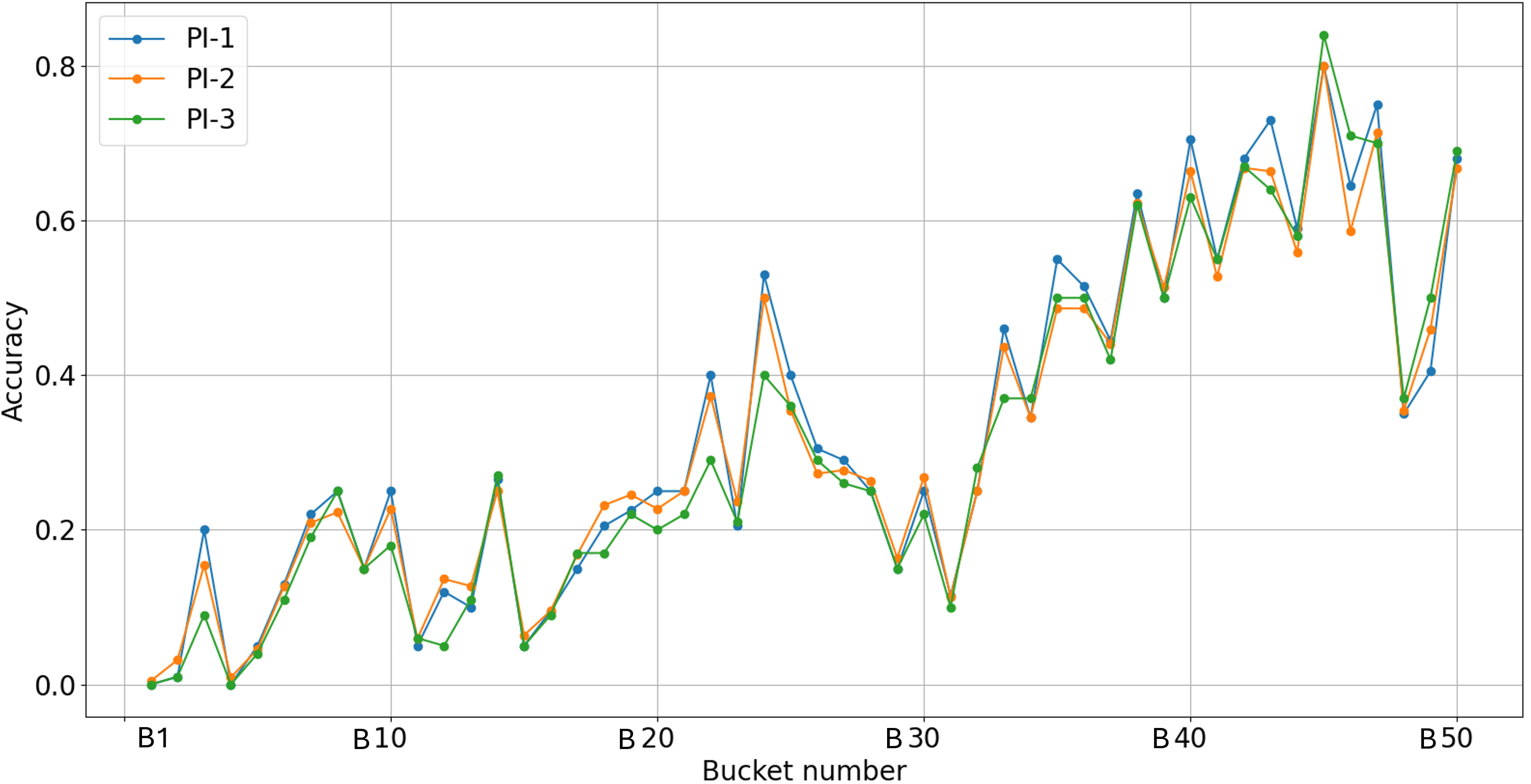}
        \caption{Accuracy ($\%$) per bucket - \gptt for ICD}
        \label{fig:PI2-ICD-GP3}
    \end{subfigure}
    \begin{subfigure}{0.49\textwidth}
        \centering
        \includegraphics[width=1.0\textwidth]{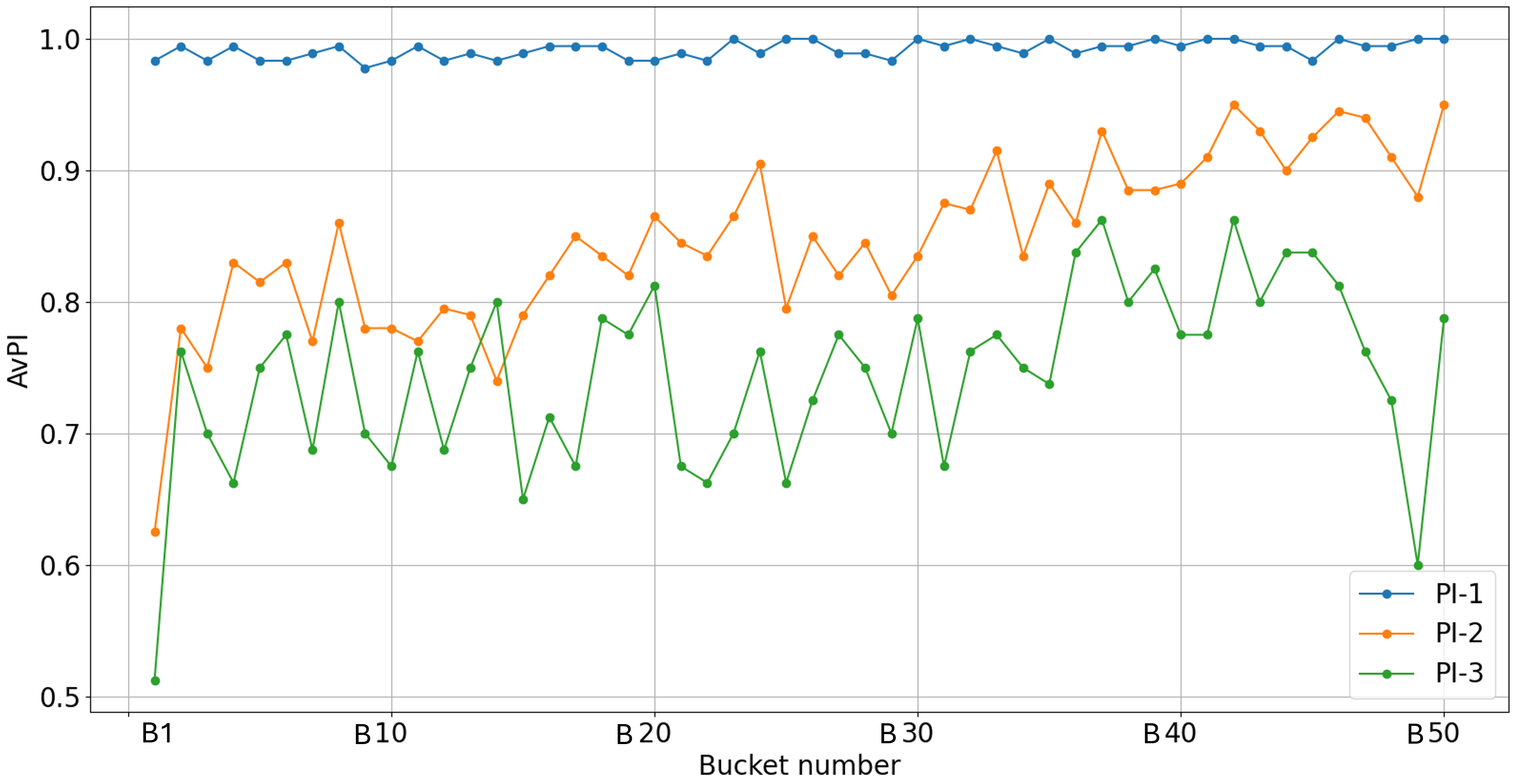}
        \caption{AvPI per bucket - GEMINI for ICD}
        \label{fig:PI-ICD-GEM}
    \end{subfigure}
    \hfill
    \begin{subfigure}{0.49\textwidth}    
        \centering
        \includegraphics[width=1.0\textwidth]{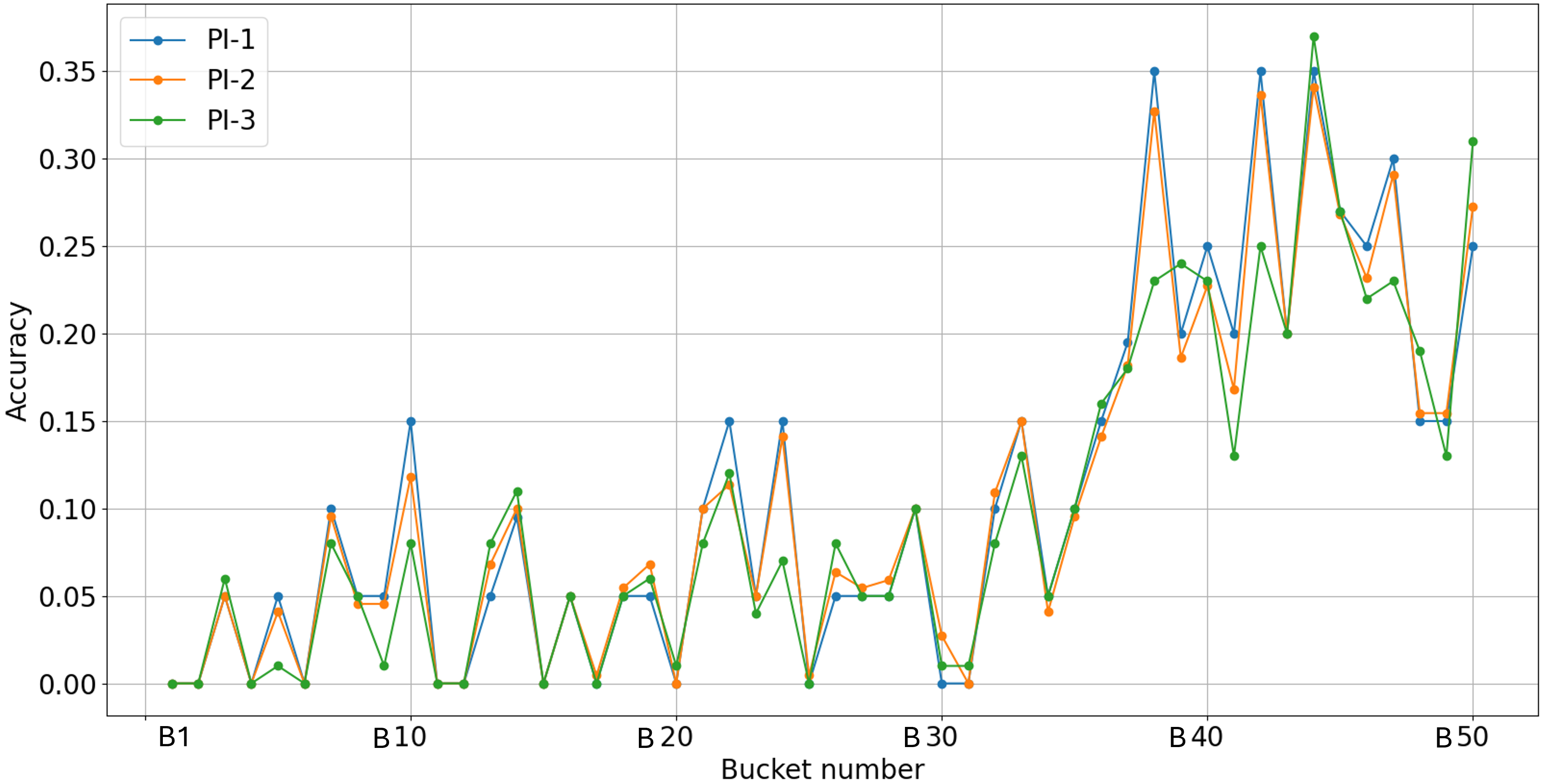}
        \caption{Accuracy ($\%$) per bucket - GEMINI for ICD}
        \label{fig:PI2-ICD-GEM}
    \end{subfigure}
    \begin{subfigure}{0.49\textwidth}
        \centering
        \includegraphics[width=1.0\textwidth]{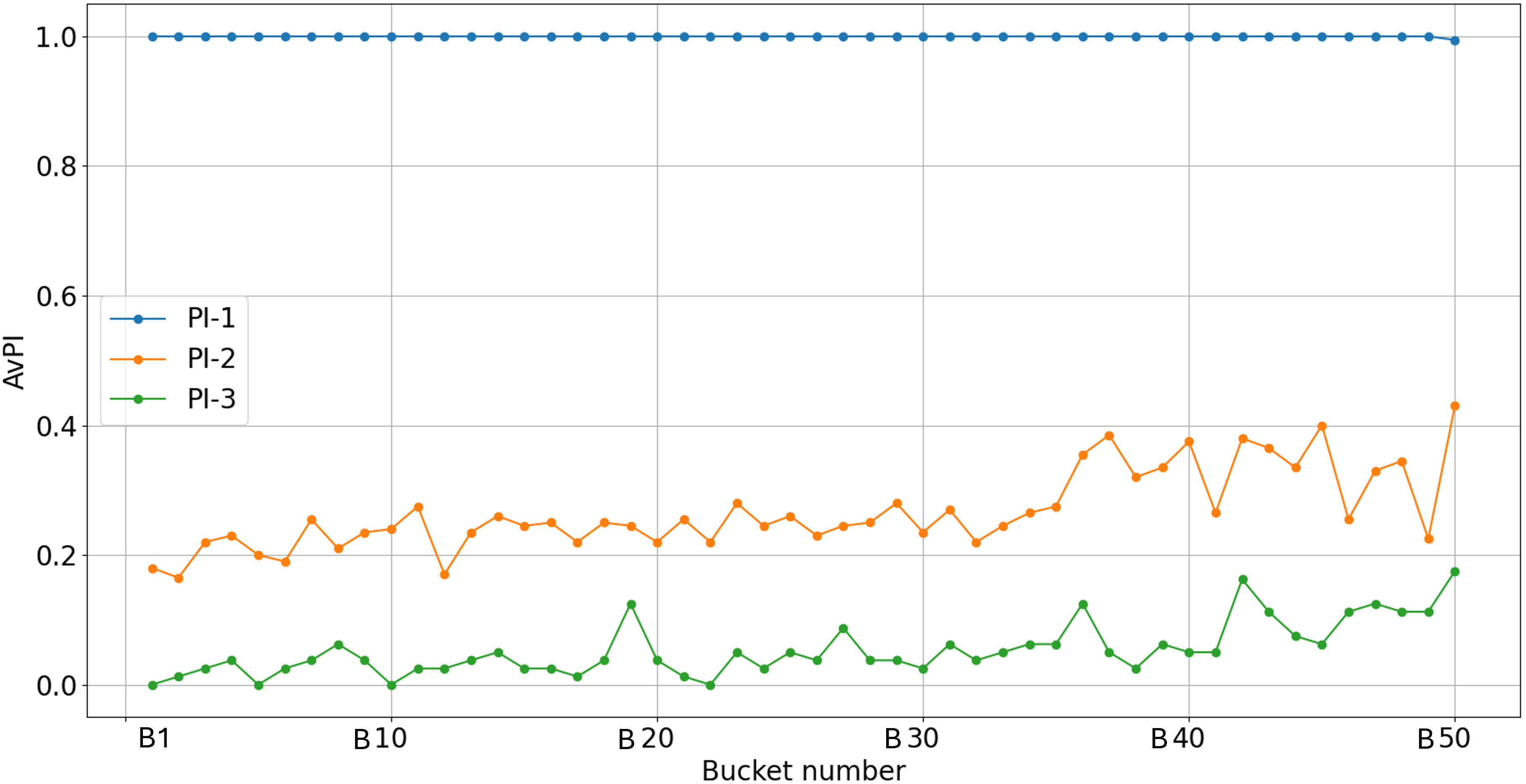}
        \caption{AvPI per bucket - Pythia for ICD}
        \label{fig:PI-ICD-PYT}
    \end{subfigure}
    \hfill
    \begin{subfigure}{0.49\textwidth}    
        \centering
        \includegraphics[width=1.0\textwidth]{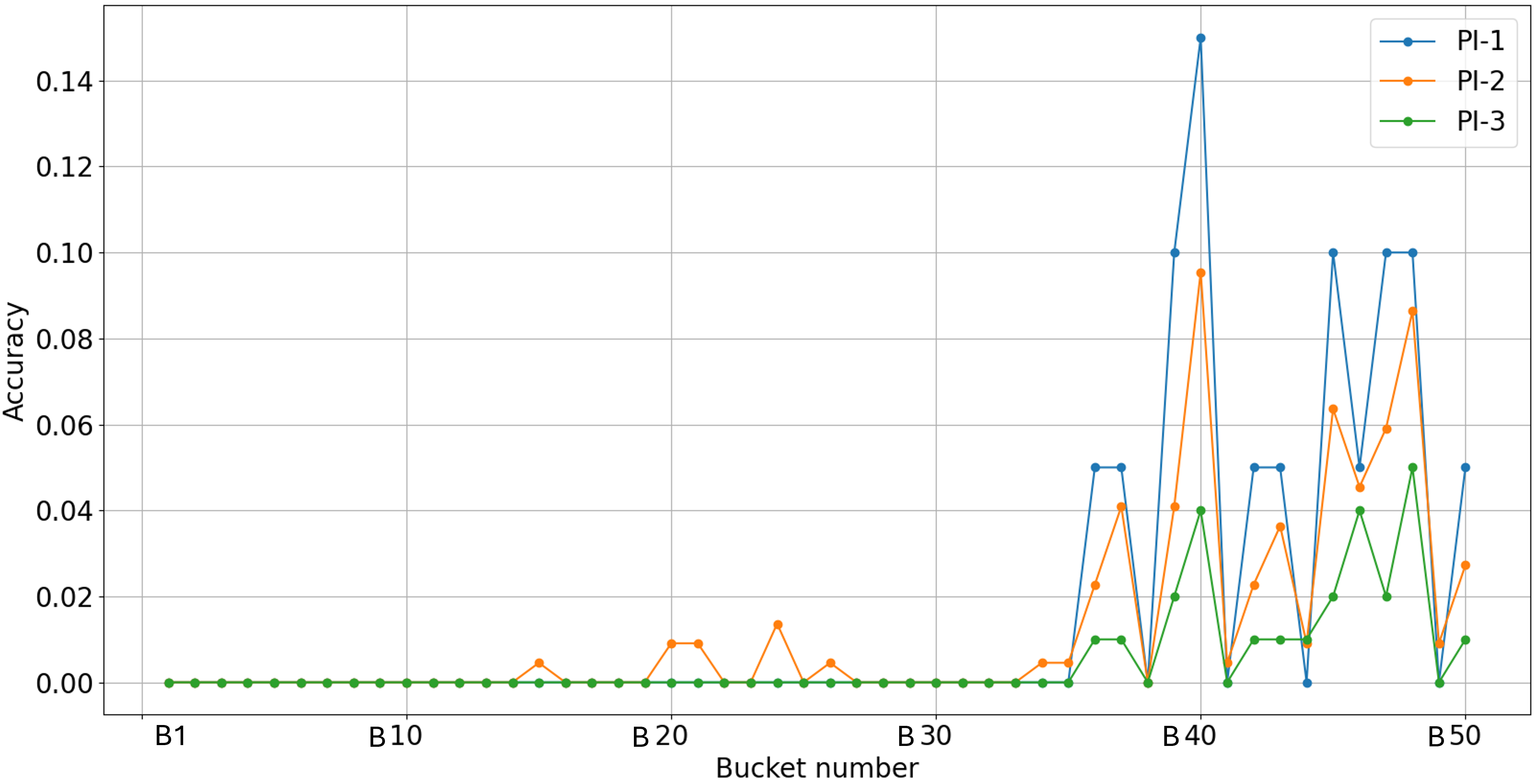}
        \caption{Accuracy ($\%$) per bucket - Pythia for ICD}
        \label{fig:PI2-ICD-PYT}
    \end{subfigure}
    \caption{
        Variation of AvPI (left) and accuracy (right) on the different buckets of ICD-10 when applying the PI-1, PI-2, and PI-3 invariance strategies to \gptt (Subfigures \ref{fig:PI-GO-GP3}-\ref{fig:PI2-GO-GP3}), \gemini (Subfigures \ref{fig:PI-ICD-GEM}-\ref{fig:PI2-ICD-GEM}) and \pythia (Subfigures \ref{fig:PI-ICD-PYT}-\ref{fig:PI2-ICD-PYT})
        }    
        \label{fig:RQ3-per-bucket-ICD}
\end{figure}

Figure~\ref{fig:RQ3-per-bucket-ICD} shows instead the trend of $AvPI$ (left) and accuracy (right) for \gptt, \gemini, and \pythia on ICD-10. The Spearman's rank coefficient between AvPI and accuracy is high (moderate to very strong correlation) for both PI-2 and PI-3 for \gptt ($.874$ and $.774$, respectively), for \gemini ($.700$ and $.535$, respectively), and for \pythia ($.613$ and $.609$, respectively). The Permutation test statistically confirms all these correlation values ($p<.05$). No correlation is observed for PI-1 for \gptt, \gemini, and \pythia.


\subsection{Discussion} \label{sub:discussionsrq3}

Figure \ref{fig:PI-GO-GP3} shows that for \gptt on the Gene Ontology, for frequently occurring concepts, i.e., those belonging to the very last buckets, all the methods achieve an average prediction invariance (AvPI) score close to 1; that is, the model almost always predicts the same ID for a given concept label, and the prediction is almost always the correct one (accuracy close to 1 as well). For infrequently occurring concepts, i.e., those belonging to the first buckets, while PI-1 achieves an AvPI score close to 1, the application of PI-2 and PI-3 results in a higher variability of the model predictions, i.e., lower AvPI score, which is more substantial when the prompt is repeated with different temperature levels (PI-2). Interestingly, in bucket B1, all methods return an average accuracy of zero, meaning that even if the methods tried different ID predictions among the various repetitions, none of them was the correct one. The increasing trend for both PI-2 and PI-3, which strongly correlates with the accuracy in Figure~\ref{fig:PI2-GO-GP3}, indicates that the more a concept is repeated on the Web (and therefore potentially in the model's training material), the more invariant the model predictions are when solicited with PI-2 and PI-3 methods. In contrast, the PI-1 method is not particularly effective for assessing correctly memorized information, as the model tends to return the same answers consistently, regardless of their accuracy. Similar trends on the Gene Ontology can also be observed for \gemini (Figures \ref{fig:PI-GO-GEM} and \ref{fig:PI2-GO-GEM}) and \pythia (Figures \ref{fig:PI-GO-PYT} and \ref{fig:PI2-GO-PYT}), although for the latter the AvPI vs accuracy correlation is weaker for PI-3 than PI-2: this may be explained by the fact that \pythia was primarily trained on English material, and therefore, a method based on multilingualism might not be optimal.

For ICD-10, similar conclusions can be reached for \gptt: the plots for AvPI (Figure~\ref{fig:PI-ICD-GP3}) and accuracy (Figure~\ref{fig:PI2-ICD-GP3}) have similar increasing trends, confirmed 
by the high correlation for both PI-2 and PI-3. Again, as for the Gene Ontology, the PI-1 method is not particularly effective for assessing correctly memorized information. \gemini has similar trends on ICD-10 as on the Gene Ontology, with moderate/high correlation for PI-2, while a less marked correlation is reported for PI-3. This may be explained by its lower knowledge of ICD-10 compared to the Gene Ontology, as noted by the lower accuracy scores in Figure~\ref{fig:dataset_occurrences_combined_main}, especially for high-frequent concepts.
Similar conclusions can also be reached for \pythia even if it presents lower AvPI variations and lower accuracy values. This is likely due to the overall lower performance of \pythia on ICD-10 than the Gene Ontology, as shown in Figure~\ref{fig:dataset_occurrences_combined_main}.

\vspace{1em}
\noindent
\textbf{Answer to RQ3:}  our results indicate that when an LLM has some knowledge of a given ontological resource\,---\, like \gptt, \gemini, and partly \pythia of the Gene Ontology, and \gptt and partly \gemini of ICD-10\,---\, looking at the variation of the predictions when invoked multiple times with the same prompt but varying temperature level (PI-2) or language (PI-3) may provide hints into the extent to which that information is memorized in the model:
if the model (almost) always predicts the same information (AvPI close to 1), it is potentially because it has seen that information many times in the training material, and thus has memorized it. Furthermore, this prompt-invariant answer is likely correct, as suggested by the (moderate to very strong) correlation between AvPI and accuracy. Conversely, if the model tends to predict different information (AvPI heading toward 0) when varying the prompt, it may be because the LLM has not memorized that concept due to its scarce occurrence in the training data.

\section{Conclusions}
\label{sec:conclusions}

In this paper, we investigate to what extent LLMs have acquired, i.e., correctly memorized, the vocabulary of concept IDs and labels from various publicly available ontologies. 
Our experiments, which span several LLMs (namely, \pythia, \gemini, \gptt and \gptf) and ontological resources (namely, Gene Ontology, Uberon, Wikidata, ICD-10) show that only a tiny fraction of concepts is memorized. Among the LLMs analyzed, \gptf is the one that obtains the highest performance.
While trying to answer the question of why some instances are correctly memorized and others not, we found for all the LLMs a substantial correlation between the popularity of the concept (i.e., the number of concept occurrences on the Web) and the correctness of the prediction. This correlation may be ascribed to the fact that the resources used for training the models likely include substantial textual material from the Web. This is indeed in line with previous research, which has shown that LLMs often achieve high performance across multiple benchmarks, but also that such performance could be attributed to potential data contamination during training and thus memorization of gold annotations \cite{deng2023investigating}.
Moreover, in our experiments, we show that the more a concept is widespread on the Web and thus likely seen in the LLMs' training data, the more the wrong prediction is closely related, in terms of Jaccard similarity, to the gold-standard one.
Finally, we proposed three different strategies for computing a prediction invariance metric for estimating the correct memorization of concepts in LLMs, showing that the invariance of the model output to the language of the prompt or the configured model temperature can be used as evidence of information memorization. 
{\modifiedblue{
Overall, our findings indicate that while LLMs acquire some structured knowledge from its presence on the Web, similar to other forms of knowledge, they also exhibit a significant tendency to ``dream'' (i.e., hallucinate) about them. This behavior aligns with their overarching objective of providing responses at all costs, prioritizing assistive interaction over factual accuracy.
}}

%

{\modifiedblue{
\section{Limitations and Future work}
We acknowledge that this study has some limitations.

First, our analysis was conducted on a limited set of LLMs, each varying in size and popularity. While models such as \gptf and \gptt are widely used, others like \pythia are comparatively less popular, which may influence the generalizability of our findings. A broader evaluation of LLMs across different architectures and sizes would provide deeper insights into how ontology memorization changes, also in relation to model advancements over time.

Second, our investigation focused exclusively on the memorization of ID-label associations in ontologies, without analyzing their structural and relational content. Exploring more complex ontological relationships, such as subsumption/generalization, could provide deeper insights into how LLMs internalize structured semantic knowledge.

Third, while we analyzed memorization patterns, we did not investigate practical interventions to mitigate hallucinations, leaving open the question of how to enhance LLMs’ reliability when dealing with structured knowledge. A promising direction for future work is exploring techniques to reduce hallucinations in LLMs when handling ontologies, potentially improving their ability to generate more accurate and reliable responses.

Moreover, this study opens several promising directions for future research. Future work could focus on refining prompting strategies to better assess how different formulations influence ID-label retrieval. Additionally, developing new metrics that complement those proposed in this study could provide a more nuanced understanding of memorization patterns. Finally, examining the impact of training data composition in a controlled setting\,---\,using models with transparent pre-training corpora, such as OLMO~\cite{bib:olmo}\,---\,could offer valuable insights into how exposure to structured data shapes LLM memory.

Our study highlights the limitations of LLMs in retaining structured knowledge and underscores the need for further research on mitigating hallucinations and enhancing ontology-aware learning. We hope this work fosters further exploration into the interaction between LLMs and structured knowledge, as addressing these challenges will ultimately improve their ability to process structured data (e.g., for answer validation and enhancing factual accuracy) while reducing unintended memorization artifacts.
}}

\section*{Data Availability}
We publicly release the resources from this study to support future research on the interaction between LLMs and structured knowledge. The code and dataset used in our experiments are available at: \\
\url{https://github.com/marcobombieri/do-LLM-dream-of-ontologies}



\begin{thebibliography}{53}


\ifx \showCODEN    \undefined \def \showCODEN     #1{\unskip}     \fi
\ifx \showDOI      \undefined \def \showDOI       #1{#1}\fi
\ifx \showISBNx    \undefined \def \showISBNx     #1{\unskip}     \fi
\ifx \showISBNxiii \undefined \def \showISBNxiii  #1{\unskip}     \fi
\ifx \showISSN     \undefined \def \showISSN      #1{\unskip}     \fi
\ifx \showLCCN     \undefined \def \showLCCN      #1{\unskip}     \fi
\ifx \shownote     \undefined \def \shownote      #1{#1}          \fi
\ifx \showarticletitle \undefined \def \showarticletitle #1{#1}   \fi
\ifx \showURL      \undefined \def \showURL       {\relax}        \fi
\providecommand\bibfield[2]{#2}
\providecommand\bibinfo[2]{#2}
\providecommand\natexlab[1]{#1}
\providecommand\showeprint[2][]{arXiv:#2}

\bibitem[Ashburner et~al\mbox{.}(2000)]%
        {geneontology}
\bibfield{author}{\bibinfo{person}{Michael Ashburner}, \bibinfo{person}{Catherine Ball}, \bibinfo{person}{Judith Blake}, \bibinfo{person}{David Botstein}, \bibinfo{person}{Heather Butler}, \bibinfo{person}{Michael Cherry}, \bibinfo{person}{Allan Davis}, \bibinfo{person}{Kara Dolinski}, \bibinfo{person}{Selina Dwight}, \bibinfo{person}{Janan Eppig}, \bibinfo{person}{Midori Harris}, \bibinfo{person}{David Hill}, \bibinfo{person}{Laurie Issel-Tarver}, \bibinfo{person}{Andrew Kasarskis}, \bibinfo{person}{Suzanna Lewis}, \bibinfo{person}{John Matese}, \bibinfo{person}{Joel Richardson}, \bibinfo{person}{Martin Ringwald}, \bibinfo{person}{Gerald Rubin}, {and} \bibinfo{person}{Gavin Sherlock}.} \bibinfo{year}{2000}\natexlab{}.
\newblock \showarticletitle{Gene ontology: tool for the unification of biology. The Gene Ontology Consortium}.
\newblock \bibinfo{journal}{\emph{Nature Genetics}} \bibinfo{volume}{25}, \bibinfo{number}{1} (\bibinfo{date}{May} \bibinfo{year}{2000}), \bibinfo{pages}{25--29}.
\newblock
\urldef\tempurl%
\url{https://doi.org/10.1038/75556}
\showDOI{\tempurl}


\bibitem[Biderman et~al\mbox{.}(2023a)]%
        {bib:predictable-memorization-2023}
\bibfield{author}{\bibinfo{person}{Stella Biderman}, \bibinfo{person}{Usvsn~Sai Prashanth}, \bibinfo{person}{Lintang Sutawika}, \bibinfo{person}{Hailey Schoelkopf}, \bibinfo{person}{Quentin Anthony}, \bibinfo{person}{Shivanshu Purohit}, {and} \bibinfo{person}{Edward Raff}.} \bibinfo{year}{2023}\natexlab{a}.
\newblock \showarticletitle{Emergent and predictable memorization in large language models}. In \bibinfo{booktitle}{\emph{Advances in Neural Information Processing Systems}}, Vol.~\bibinfo{volume}{36}. \bibinfo{publisher}{Curran Associates Inc.}, \bibinfo{address}{Red Hook, NY, USA}, \bibinfo{pages}{28072--28090}.
\newblock
\urldef\tempurl%
\url{https://proceedings.neurips.cc/paper_files/paper/2023/file/59404fb89d6194641c69ae99ecdf8f6d-Paper-Conference.pdf}
\showURL{%
\tempurl}


\bibitem[Biderman et~al\mbox{.}(2023b)]%
        {Pythia2023}
\bibfield{author}{\bibinfo{person}{Stella Biderman}, \bibinfo{person}{Hailey Schoelkopf}, \bibinfo{person}{Quentin~Gregory Anthony}, \bibinfo{person}{Herbie Bradley}, \bibinfo{person}{Kyle O'Brien}, \bibinfo{person}{Eric Hallahan}, \bibinfo{person}{Mohammad~Aflah Khan}, \bibinfo{person}{Shivanshu Purohit}, \bibinfo{person}{USVSN~Sai Prashanth}, \bibinfo{person}{Edward Raff}, \bibinfo{person}{Aviya Skowron}, \bibinfo{person}{Lintang Sutawika}, {and} \bibinfo{person}{Oskar van~der Wal}.} \bibinfo{year}{2023}\natexlab{b}.
\newblock \showarticletitle{Pythia: {A} Suite for Analyzing Large Language Models Across Training and Scaling}. In \bibinfo{booktitle}{\emph{Proceedings of the 40th International Conference on Machine Learning}}. \bibinfo{publisher}{{PMLR}}, \bibinfo{address}{Honolulu, Hawaii, {USA}}, \bibinfo{pages}{2397--2430}.
\newblock
\urldef\tempurl%
\url{https://proceedings.mlr.press/v202/biderman23a.html}
\showURL{%
\tempurl}


\bibitem[{BigScience Workshop}(2023)]%
        {workshop2023bloom}
\bibfield{author}{\bibinfo{person}{{BigScience Workshop}}.} \bibinfo{year}{2023}\natexlab{}.
\newblock \bibinfo{title}{BLOOM: A 176B-Parameter Open-Access Multilingual Language Model}.
\newblock
\newblock
\showeprint[arxiv]{2211.05100}~[cs.CL]
\urldef\tempurl%
\url{https://arxiv.org/abs/2211.05100}
\showURL{%
\tempurl}


\bibitem[Boer et~al\mbox{.}(2024)]%
        {triplet}
\bibfield{author}{\bibinfo{person}{Derian Boer}, \bibinfo{person}{Fabian Koch}, {and} \bibinfo{person}{Stefan Kramer}.} \bibinfo{year}{2024}\natexlab{}.
\newblock \bibinfo{title}{Harnessing the Power of Semi-Structured Knowledge and LLMs with Triplet-Based Prefiltering for Question Answering}.
\newblock
\newblock
\showeprint[arxiv]{2409.00861}~[cs.CL]
\urldef\tempurl%
\url{https://arxiv.org/abs/2409.00861}
\showURL{%
\tempurl}


\bibitem[Brown et~al\mbox{.}(2021)]%
        {memorization-learning2}
\bibfield{author}{\bibinfo{person}{Gavin Brown}, \bibinfo{person}{Mark Bun}, \bibinfo{person}{Vitaly Feldman}, \bibinfo{person}{Adam Smith}, {and} \bibinfo{person}{Kunal Talwar}.} \bibinfo{year}{2021}\natexlab{}.
\newblock \showarticletitle{When is memorization of irrelevant training data necessary for high-accuracy learning?}. In \bibinfo{booktitle}{\emph{Proceedings of the 53rd Annual ACM SIGACT Symposium on Theory of Computing}} (Virtual, Italy). \bibinfo{publisher}{Association for Computing Machinery}, \bibinfo{address}{New York, NY, USA}, \bibinfo{pages}{123–132}.
\newblock
\showISBNx{9781450380539}
\urldef\tempurl%
\url{https://doi.org/10.1145/3406325.3451131}
\showDOI{\tempurl}


\bibitem[Brown et~al\mbox{.}(2020)]%
        {gpt3}
\bibfield{author}{\bibinfo{person}{Tom Brown}, \bibinfo{person}{Benjamin Mann}, \bibinfo{person}{Nick Ryder}, \bibinfo{person}{Melanie Subbiah}, \bibinfo{person}{Jared~D Kaplan}, \bibinfo{person}{Prafulla Dhariwal}, \bibinfo{person}{Arvind Neelakantan}, \bibinfo{person}{Pranav Shyam}, \bibinfo{person}{Girish Sastry}, \bibinfo{person}{Amanda Askell}, \bibinfo{person}{Sandhini Agarwal}, \bibinfo{person}{Ariel Herbert-Voss}, \bibinfo{person}{Gretchen Krueger}, \bibinfo{person}{Tom Henighan}, \bibinfo{person}{Rewon Child}, \bibinfo{person}{Aditya Ramesh}, \bibinfo{person}{Daniel Ziegler}, \bibinfo{person}{Jeffrey Wu}, \bibinfo{person}{Clemens Winter}, \bibinfo{person}{Chris Hesse}, \bibinfo{person}{Mark Chen}, \bibinfo{person}{Eric Sigler}, \bibinfo{person}{Mateusz Litwin}, \bibinfo{person}{Scott Gray}, \bibinfo{person}{Benjamin Chess}, \bibinfo{person}{Jack Clark}, \bibinfo{person}{Christopher Berner}, \bibinfo{person}{Sam McCandlish}, \bibinfo{person}{Alec Radford}, \bibinfo{person}{Ilya Sutskever}, {and}
  \bibinfo{person}{Dario Amodei}.} \bibinfo{year}{2020}\natexlab{}.
\newblock \showarticletitle{Language Models are Few-Shot Learners}. In \bibinfo{booktitle}{\emph{Advances in Neural Information Processing Systems}}, Vol.~\bibinfo{volume}{33}. \bibinfo{publisher}{Curran Associates, Inc.}, \bibinfo{address}{Virtual}, \bibinfo{pages}{1877--1901}.
\newblock
\urldef\tempurl%
\url{https://proceedings.neurips.cc/paper_files/paper/2020/file/1457c0d6bfcb4967418bfb8ac142f64a-Paper.pdf}
\showURL{%
\tempurl}


\bibitem[Carlini et~al\mbox{.}(2023)]%
        {Carlini2023}
\bibfield{author}{\bibinfo{person}{Nicholas Carlini}, \bibinfo{person}{Daphne Ippolito}, \bibinfo{person}{Matthew Jagielski}, \bibinfo{person}{Katherine Lee}, \bibinfo{person}{Florian Tramer}, {and} \bibinfo{person}{Chiyuan Zhang}.} \bibinfo{year}{2023}\natexlab{}.
\newblock \bibinfo{title}{Quantifying Memorization Across Neural Language Models}.
\newblock
\newblock
\showeprint[arxiv]{2202.07646}~[cs.LG]
\urldef\tempurl%
\url{https://arxiv.org/abs/2202.07646}
\showURL{%
\tempurl}


\bibitem[Carlini et~al\mbox{.}(2019)]%
        {bib-unintended-memorization-2019}
\bibfield{author}{\bibinfo{person}{Nicholas Carlini}, \bibinfo{person}{Chang Liu}, \bibinfo{person}{{\'{U}}lfar Erlingsson}, \bibinfo{person}{Jernej Kos}, {and} \bibinfo{person}{Dawn Song}.} \bibinfo{year}{2019}\natexlab{}.
\newblock \showarticletitle{The Secret Sharer: Evaluating and Testing Unintended Memorization in Neural Networks}. In \bibinfo{booktitle}{\emph{28th {USENIX} Security Symposium}}. \bibinfo{publisher}{{USENIX} Association}, \bibinfo{address}{Santa Clara, CA, USA}, \bibinfo{pages}{267--284}.
\newblock
\urldef\tempurl%
\url{https://www.usenix.org/conference/usenixsecurity19/presentation/carlini}
\showURL{%
\tempurl}


\bibitem[Chang et~al\mbox{.}(2023)]%
        {Bamman2023}
\bibfield{author}{\bibinfo{person}{Kent Chang}, \bibinfo{person}{Mackenzie Cramer}, \bibinfo{person}{Sandeep Soni}, {and} \bibinfo{person}{David Bamman}.} \bibinfo{year}{2023}\natexlab{}.
\newblock \showarticletitle{Speak, Memory: An Archaeology of Books Known to ChatGPT/GPT-4}. In \bibinfo{booktitle}{\emph{Proceedings of the 2023 Conference on Empirical Methods in Natural Language Processing}}. \bibinfo{publisher}{Association for Computational Linguistics}, \bibinfo{address}{Singapore}, \bibinfo{pages}{7312--7327}.
\newblock
\urldef\tempurl%
\url{https://doi.org/10.18653/v1/2023.emnlp-main.453}
\showDOI{\tempurl}


\bibitem[Chiang et~al\mbox{.}(2024)]%
        {chiang2024chatbotarenaopenplatform}
\bibfield{author}{\bibinfo{person}{Wei-Lin Chiang}, \bibinfo{person}{Lianmin Zheng}, \bibinfo{person}{Ying Sheng}, \bibinfo{person}{Anastasios~Nikolas Angelopoulos}, \bibinfo{person}{Tianle Li}, \bibinfo{person}{Dacheng Li}, \bibinfo{person}{Hao Zhang}, \bibinfo{person}{Banghua Zhu}, \bibinfo{person}{Michael Jordan}, \bibinfo{person}{Joseph~E. Gonzalez}, {and} \bibinfo{person}{Ion Stoica}.} \bibinfo{year}{2024}\natexlab{}.
\newblock \bibinfo{title}{Chatbot Arena: An Open Platform for Evaluating LLMs by Human Preference}.
\newblock
\newblock
\showeprint[arxiv]{2403.04132}~[cs.AI]
\urldef\tempurl%
\url{https://arxiv.org/abs/2403.04132}
\showURL{%
\tempurl}


\bibitem[Chuang et~al\mbox{.}(2024)]%
        {hallucination3}
\bibfield{author}{\bibinfo{person}{Yung{-}Sung Chuang}, \bibinfo{person}{Yujia Xie}, \bibinfo{person}{Hongyin Luo}, \bibinfo{person}{Yoon Kim}, \bibinfo{person}{James~R. Glass}, {and} \bibinfo{person}{Pengcheng He}.} \bibinfo{year}{2024}\natexlab{}.
\newblock \bibinfo{title}{DoLa: Decoding by Contrasting Layers Improves Factuality in Large Language Models}.
\newblock
\newblock
\showeprint[arxiv]{2309.03883}~[cs.CL]
\urldef\tempurl%
\url{https://arxiv.org/abs/2309.03883}
\showURL{%
\tempurl}


\bibitem[Deng et~al\mbox{.}(2024)]%
        {deng2023investigating}
\bibfield{author}{\bibinfo{person}{Chunyuan Deng}, \bibinfo{person}{Yilun Zhao}, \bibinfo{person}{Xiangru Tang}, \bibinfo{person}{Mark Gerstein}, {and} \bibinfo{person}{Arman Cohan}.} \bibinfo{year}{2024}\natexlab{}.
\newblock \bibinfo{title}{Investigating Data Contamination in Modern Benchmarks for Large Language Models}.
\newblock
\newblock
\showeprint[arxiv]{2311.09783}~[cs.CL]
\urldef\tempurl%
\url{https://arxiv.org/abs/2311.09783}
\showURL{%
\tempurl}


\bibitem[Feldman(2020)]%
        {memorization-learning}
\bibfield{author}{\bibinfo{person}{Vitaly Feldman}.} \bibinfo{year}{2020}\natexlab{}.
\newblock \showarticletitle{Does learning require memorization? a short tale about a long tail}. In \bibinfo{booktitle}{\emph{Proceedings of the 52nd Annual ACM SIGACT Symposium on Theory of Computing}} (Chicago, IL, USA). \bibinfo{publisher}{Association for Computing Machinery}, \bibinfo{address}{New York, NY, USA}, \bibinfo{pages}{954–959}.
\newblock
\showISBNx{9781450369794}
\urldef\tempurl%
\url{https://doi.org/10.1145/3357713.3384290}
\showDOI{\tempurl}


\bibitem[Frey et~al\mbox{.}(2023)]%
        {rdf}
\bibfield{author}{\bibinfo{person}{Johannes Frey}, \bibinfo{person}{Lars-Peter Meyer}, \bibinfo{person}{Natanael Arndt}, \bibinfo{person}{Felix Brei}, {and} \bibinfo{person}{Kirill Bulert}.} \bibinfo{year}{2023}\natexlab{}.
\newblock \bibinfo{title}{Benchmarking the Abilities of Large Language Models for {RDF} Knowledge Graph Creation and Comprehension: How Well Do LLMs Speak Turtle?}
\newblock
\newblock
\showeprint[arxiv]{2309.17122}~[cs.AI]
\urldef\tempurl%
\url{https://arxiv.org/abs/2309.17122}
\showURL{%
\tempurl}


\bibitem[Gao et~al\mbox{.}(2020)]%
        {thePile}
\bibfield{author}{\bibinfo{person}{Leo Gao}, \bibinfo{person}{Stella Biderman}, \bibinfo{person}{Sid Black}, \bibinfo{person}{Laurence Golding}, \bibinfo{person}{Travis Hoppe}, \bibinfo{person}{Charles Foster}, \bibinfo{person}{Jason Phang}, \bibinfo{person}{Horace He}, \bibinfo{person}{Anish Thite}, \bibinfo{person}{Noa Nabeshima}, \bibinfo{person}{Shawn Presser}, {and} \bibinfo{person}{Connor Leahy}.} \bibinfo{year}{2020}\natexlab{}.
\newblock \bibinfo{title}{The Pile: An 800GB Dataset of Diverse Text for Language Modeling}.
\newblock
\newblock
\showeprint[arxiv]{2101.00027}~[cs.CL]
\urldef\tempurl%
\url{https://arxiv.org/abs/2101.00027}
\showURL{%
\tempurl}


\bibitem[Giunchiglia and Zaihrayeu(2009)]%
        {Giunchiglia2009}
\bibfield{author}{\bibinfo{person}{Fausto Giunchiglia} {and} \bibinfo{person}{Ilya Zaihrayeu}.} \bibinfo{year}{2009}\natexlab{}.
\newblock \bibinfo{booktitle}{\emph{Lightweight Ontologies}}.
\newblock \bibinfo{publisher}{Springer US}, \bibinfo{address}{Boston, MA}, \bibinfo{pages}{1613--1619}.
\newblock
\showISBNx{978-0-387-39940-9}
\urldef\tempurl%
\url{https://doi.org/10.1007/978-0-387-39940-9_1314}
\showDOI{\tempurl}


\bibitem[Granger(1969)]%
        {bib:granger}
\bibfield{author}{\bibinfo{person}{Clive William~John Granger}.} \bibinfo{year}{1969}\natexlab{}.
\newblock \showarticletitle{{Investigating Causal Relations by Econometric Models and Cross-Spectral Methods}}.
\newblock \bibinfo{journal}{\emph{Econometrica}} \bibinfo{volume}{37}, \bibinfo{number}{3} (\bibinfo{date}{July} \bibinfo{year}{1969}), \bibinfo{pages}{424--438}.
\newblock
\urldef\tempurl%
\url{https://ideas.repec.org/a/ecm/emetrp/v37y1969i3p424-38.html}
\showURL{%
\tempurl}


\bibitem[Groeneveld et~al\mbox{.}(2024)]%
        {bib:olmo}
\bibfield{author}{\bibinfo{person}{Dirk Groeneveld}, \bibinfo{person}{Iz Beltagy}, \bibinfo{person}{Evan~Pete Walsh}, \bibinfo{person}{Akshita Bhagia}, \bibinfo{person}{Rodney Kinney}, \bibinfo{person}{Oyvind Tafjord}, \bibinfo{person}{Ananya~Harsh Jha}, \bibinfo{person}{Hamish Ivison}, \bibinfo{person}{Ian Magnusson}, \bibinfo{person}{Yizhong Wang}, \bibinfo{person}{Shane Arora}, \bibinfo{person}{David Atkinson}, \bibinfo{person}{Russell Authur}, \bibinfo{person}{Khyathi~Raghavi Chandu}, \bibinfo{person}{Arman Cohan}, \bibinfo{person}{Jennifer Dumas}, \bibinfo{person}{Yanai Elazar}, \bibinfo{person}{Yuling Gu}, \bibinfo{person}{Jack Hessel}, \bibinfo{person}{Tushar Khot}, \bibinfo{person}{William Merrill}, \bibinfo{person}{Jacob Morrison}, \bibinfo{person}{Niklas Muennighoff}, \bibinfo{person}{Aakanksha Naik}, \bibinfo{person}{Crystal Nam}, \bibinfo{person}{Matthew~E. Peters}, \bibinfo{person}{Valentina Pyatkin}, \bibinfo{person}{Abhilasha Ravichander}, \bibinfo{person}{Dustin Schwenk},
  \bibinfo{person}{Saurabh Shah}, \bibinfo{person}{Will Smith}, \bibinfo{person}{Emma Strubell}, \bibinfo{person}{Nishant Subramani}, \bibinfo{person}{Mitchell Wortsman}, \bibinfo{person}{Pradeep Dasigi}, \bibinfo{person}{Nathan Lambert}, \bibinfo{person}{Kyle Richardson}, \bibinfo{person}{Luke Zettlemoyer}, \bibinfo{person}{Jesse Dodge}, \bibinfo{person}{Kyle Lo}, \bibinfo{person}{Luca Soldaini}, \bibinfo{person}{Noah~A. Smith}, {and} \bibinfo{person}{Hannaneh Hajishirzi}.} \bibinfo{year}{2024}\natexlab{}.
\newblock \showarticletitle{OLMo: Accelerating the Science of Language Models}. In \bibinfo{booktitle}{\emph{Proceedings of the 62nd Annual Meeting of the Association for Computational Linguistics (Volume 1: Long Papers)}}. \bibinfo{publisher}{Association for Computational Linguistics}, \bibinfo{address}{Bangkok, Thailand}, \bibinfo{pages}{15789--15809}.
\newblock
\urldef\tempurl%
\url{https://doi.org/10.18653/V1/2024.ACL-LONG.841}
\showDOI{\tempurl}


\bibitem[He et~al\mbox{.}(2023)]%
        {Horrocks2023}
\bibfield{author}{\bibinfo{person}{Yuan He}, \bibinfo{person}{Jiaoyan Chen}, \bibinfo{person}{Ernesto Jim{\'{e}}nez{-}Ruiz}, \bibinfo{person}{Hang Dong}, {and} \bibinfo{person}{Ian Horrocks}.} \bibinfo{year}{2023}\natexlab{}.
\newblock \showarticletitle{Language Model Analysis for Ontology Subsumption Inference}. In \bibinfo{booktitle}{\emph{Findings of the Association for Computational Linguistics: {ACL}}}. \bibinfo{publisher}{Association for Computational Linguistics}, \bibinfo{address}{Toronto, Canada}, \bibinfo{pages}{3439--3453}.
\newblock
\urldef\tempurl%
\url{https://doi.org/10.18653/v1/2023.findings-acl.213}
\showDOI{\tempurl}


\bibitem[Hendrycks et~al\mbox{.}(2021)]%
        {hendryckstest2021}
\bibfield{author}{\bibinfo{person}{Dan Hendrycks}, \bibinfo{person}{Collin Burns}, \bibinfo{person}{Steven Basart}, \bibinfo{person}{Andy Zou}, \bibinfo{person}{Mantas Mazeika}, \bibinfo{person}{Dawn Song}, {and} \bibinfo{person}{Jacob Steinhardt}.} \bibinfo{year}{2021}\natexlab{}.
\newblock \bibinfo{title}{Measuring Massive Multitask Language Understanding}.
\newblock
\newblock
\showeprint[arxiv]{2009.03300}~[cs.CY]
\urldef\tempurl%
\url{https://arxiv.org/abs/2009.03300}
\showURL{%
\tempurl}


\bibitem[Hertling and Paulheim(2023)]%
        {Paulheim2023}
\bibfield{author}{\bibinfo{person}{Sven Hertling} {and} \bibinfo{person}{Heiko Paulheim}.} \bibinfo{year}{2023}\natexlab{}.
\newblock \showarticletitle{OLaLa: Ontology Matching with Large Language Models}. In \bibinfo{booktitle}{\emph{Proceedings of the 12th Knowledge Capture Conference}}. \bibinfo{publisher}{{ACM}}, \bibinfo{address}{Pensacola, {FL, USA}}, \bibinfo{pages}{131--139}.
\newblock
\urldef\tempurl%
\url{https://doi.org/10.1145/3587259.3627571}
\showDOI{\tempurl}


\bibitem[Huang et~al\mbox{.}(2022)]%
        {huang-etal-2022-large}
\bibfield{author}{\bibinfo{person}{Jie Huang}, \bibinfo{person}{Hanyin Shao}, {and} \bibinfo{person}{Kevin Chen-Chuan Chang}.} \bibinfo{year}{2022}\natexlab{}.
\newblock \showarticletitle{Are Large Pre-Trained Language Models Leaking Your Personal Information?}. In \bibinfo{booktitle}{\emph{Findings of the Association for Computational Linguistics: EMNLP}}. \bibinfo{publisher}{Association for Computational Linguistics}, \bibinfo{address}{Abu Dhabi, United Arab Emirates}, \bibinfo{pages}{2038--2047}.
\newblock
\urldef\tempurl%
\url{https://doi.org/10.18653/v1/2022.findings-emnlp.148}
\showDOI{\tempurl}


\bibitem[Huang et~al\mbox{.}(2025)]%
        {bib:hallucination-new}
\bibfield{author}{\bibinfo{person}{Lei Huang}, \bibinfo{person}{Weijiang Yu}, \bibinfo{person}{Weitao Ma}, \bibinfo{person}{Weihong Zhong}, \bibinfo{person}{Zhangyin Feng}, \bibinfo{person}{Haotian Wang}, \bibinfo{person}{Qianglong Chen}, \bibinfo{person}{Weihua Peng}, \bibinfo{person}{Xiaocheng Feng}, \bibinfo{person}{Bing Qin}, {and} \bibinfo{person}{Ting Liu}.} \bibinfo{year}{2025}\natexlab{}.
\newblock \showarticletitle{A Survey on Hallucination in Large Language Models: Principles, Taxonomy, Challenges, and Open Questions}.
\newblock \bibinfo{journal}{\emph{ACM Transactions on Information Systems}} \bibinfo{volume}{43}, \bibinfo{number}{2}, Article \bibinfo{articleno}{42} (\bibinfo{date}{Jan.} \bibinfo{year}{2025}), \bibinfo{numpages}{55}~pages.
\newblock
\showISSN{1046-8188}
\urldef\tempurl%
\url{https://doi.org/10.1145/3703155}
\showDOI{\tempurl}


\bibitem[Ishihara(2023)]%
        {Ishihara2023}
\bibfield{author}{\bibinfo{person}{Shotaro Ishihara}.} \bibinfo{year}{2023}\natexlab{}.
\newblock \bibinfo{title}{Training Data Extraction From Pre-trained Language Models: A Survey}.
\newblock
\newblock
\showeprint[arxiv]{2305.16157}~[cs.CL]
\urldef\tempurl%
\url{https://arxiv.org/abs/2305.16157}
\showURL{%
\tempurl}


\bibitem[Jaccard(1901)]%
        {Jaccard}
\bibfield{author}{\bibinfo{person}{Paul Jaccard}.} \bibinfo{year}{1901}\natexlab{}.
\newblock \showarticletitle{Étude comparative de la distribution florale dans une portion des Alpes et des Jura}.
\newblock \bibinfo{journal}{\emph{Bulletin de la Société Vaudoise des Sciences Naturelles}}  \bibinfo{volume}{37} (\bibinfo{year}{1901}), \bibinfo{pages}{547--579}.
\newblock


\bibitem[Kahana et~al\mbox{.}(2022)]%
        {kahana_diamond_aka_2022}
\bibfield{author}{\bibinfo{person}{Michael~J. Kahana}, \bibinfo{person}{Nicholas~B. Diamond}, {and} \bibinfo{person}{Akram Aka}.} \bibinfo{year}{2022}\natexlab{}.
\newblock \bibinfo{title}{Laws of Human Memory}.
\newblock \bibinfo{howpublished}{Preprint}.
\newblock
\urldef\tempurl%
\url{https://doi.org/10.31234/osf.io/aczu9}
\showURL{%
\tempurl}


\bibitem[Kandpal et~al\mbox{.}(2022)]%
        {bib-remove-memorization2}
\bibfield{author}{\bibinfo{person}{Nikhil Kandpal}, \bibinfo{person}{Eric Wallace}, {and} \bibinfo{person}{Colin Raffel}.} \bibinfo{year}{2022}\natexlab{}.
\newblock \showarticletitle{Deduplicating Training Data Mitigates Privacy Risks in Language Models}. In \bibinfo{booktitle}{\emph{Proceedings of the 39th International Conference on Machine Learning}}. \bibinfo{publisher}{{PMLR}}, \bibinfo{address}{Baltimore, Maryland, {USA}}, \bibinfo{pages}{10697--10707}.
\newblock
\urldef\tempurl%
\url{https://proceedings.mlr.press/v162/kandpal22a.html}
\showURL{%
\tempurl}


\bibitem[Kilgarriff(2007)]%
        {kilgarriff-2007-last}
\bibfield{author}{\bibinfo{person}{Adam Kilgarriff}.} \bibinfo{year}{2007}\natexlab{}.
\newblock \showarticletitle{Last Words: Googleology is Bad Science}.
\newblock \bibinfo{journal}{\emph{Computational Linguistics}} \bibinfo{volume}{33}, \bibinfo{number}{1} (\bibinfo{year}{2007}), \bibinfo{pages}{147--151}.
\newblock
\urldef\tempurl%
\url{https://doi.org/10.1162/coli.2007.33.1.147}
\showDOI{\tempurl}


\bibitem[Koeling et~al\mbox{.}(2005)]%
        {koeling-etal-2005-domain}
\bibfield{author}{\bibinfo{person}{Rob Koeling}, \bibinfo{person}{Diana McCarthy}, {and} \bibinfo{person}{John Carroll}.} \bibinfo{year}{2005}\natexlab{}.
\newblock \showarticletitle{Domain-Specific Sense Distributions and Predominant Sense Acquisition}. In \bibinfo{booktitle}{\emph{Proceedings of Human Language Technology Conference and Conference on Empirical Methods in Natural Language Processing}}. \bibinfo{publisher}{Association for Computational Linguistics}, \bibinfo{address}{Vancouver, British Columbia, Canada}, \bibinfo{pages}{419--426}.
\newblock
\urldef\tempurl%
\url{https://aclanthology.org/H05-1053}
\showURL{%
\tempurl}


\bibitem[Lee et~al\mbox{.}(2022)]%
        {bib-remove-memorization1}
\bibfield{author}{\bibinfo{person}{Katherine Lee}, \bibinfo{person}{Daphne Ippolito}, \bibinfo{person}{Andrew Nystrom}, \bibinfo{person}{Chiyuan Zhang}, \bibinfo{person}{Douglas Eck}, \bibinfo{person}{Chris Callison{-}Burch}, {and} \bibinfo{person}{Nicholas Carlini}.} \bibinfo{year}{2022}\natexlab{}.
\newblock \showarticletitle{Deduplicating Training Data Makes Language Models Better}. In \bibinfo{booktitle}{\emph{Proceedings of the 60th Annual Meeting of the Association for Computational Linguistics (Volume 1: Long Papers)}}. \bibinfo{publisher}{Association for Computational Linguistics}, \bibinfo{address}{Dublin, Ireland}, \bibinfo{pages}{8424--8445}.
\newblock
\urldef\tempurl%
\url{https://doi.org/10.18653/v1/2022.acl-long.577}
\showDOI{\tempurl}


\bibitem[Lehman et~al\mbox{.}(2021)]%
        {lehman-etal-2021-bert}
\bibfield{author}{\bibinfo{person}{Eric Lehman}, \bibinfo{person}{Sarthak Jain}, \bibinfo{person}{Karl Pichotta}, \bibinfo{person}{Yoav Goldberg}, {and} \bibinfo{person}{Byron Wallace}.} \bibinfo{year}{2021}\natexlab{}.
\newblock \showarticletitle{Does {BERT} Pretrained on Clinical Notes Reveal Sensitive Data?}. In \bibinfo{booktitle}{\emph{Proceedings of the 2021 Conference of the North American Chapter of the Association for Computational Linguistics: Human Language Technologies}}. \bibinfo{publisher}{Association for Computational Linguistics}, \bibinfo{address}{Online}, \bibinfo{pages}{946--959}.
\newblock
\urldef\tempurl%
\url{https://doi.org/10.18653/v1/2021.naacl-main.73}
\showDOI{\tempurl}


\bibitem[Levenshtein(1966)]%
        {Levenshtein}
\bibfield{author}{\bibinfo{person}{Vladimir~Iosifovich Levenshtein}.} \bibinfo{year}{1966}\natexlab{}.
\newblock \showarticletitle{Binary codes capable of correcting deletions, insertions and reversals.}
\newblock \bibinfo{journal}{\emph{Soviet Physics Doklady}} \bibinfo{volume}{10}, \bibinfo{number}{8} (\bibinfo{date}{feb} \bibinfo{year}{1966}), \bibinfo{pages}{707--710}.
\newblock
\newblock
\shownote{Doklady Akademii Nauk SSSR, V163 No4 845-848 1965}.


\bibitem[Liu et~al\mbox{.}(2024)]%
        {bib-pretraining-detection-2024}
\bibfield{author}{\bibinfo{person}{Zhenhua Liu}, \bibinfo{person}{Tong Zhu}, \bibinfo{person}{Chuanyuan Tan}, \bibinfo{person}{Bing Liu}, \bibinfo{person}{Haonan Lu}, {and} \bibinfo{person}{Wenliang Chen}.} \bibinfo{year}{2024}\natexlab{}.
\newblock \showarticletitle{Probing Language Models for Pre-training Data Detection}. In \bibinfo{booktitle}{\emph{Proceedings of the 62nd Annual Meeting of the Association for Computational Linguistics (Volume 1: Long Papers)}}. \bibinfo{publisher}{Association for Computational Linguistics}, \bibinfo{address}{Bangkok, Thailand}, \bibinfo{pages}{1576--1587}.
\newblock
\urldef\tempurl%
\url{https://doi.org/10.18653/v1/2024.acl-long.86}
\showDOI{\tempurl}


\bibitem[Lu et~al\mbox{.}(2017)]%
        {Lu_Su_White_2017}
\bibfield{author}{\bibinfo{person}{Xun Lu}, \bibinfo{person}{Liangjun Su}, {and} \bibinfo{person}{Halbert White}.} \bibinfo{year}{2017}\natexlab{}.
\newblock \showarticletitle{Granger causality and structural causality in cross-section and panel data}.
\newblock \bibinfo{journal}{\emph{Econometric Theory}} \bibinfo{volume}{33}, \bibinfo{number}{2} (\bibinfo{year}{2017}), \bibinfo{pages}{263–291}.
\newblock
\urldef\tempurl%
\url{https://www.jstor.org/stable/26173622}
\showURL{%
\tempurl}


\bibitem[Lukas et~al\mbox{.}(2023)]%
        {bib-data-leakage-2023}
\bibfield{author}{\bibinfo{person}{Nils Lukas}, \bibinfo{person}{Ahmed Salem}, \bibinfo{person}{Robert Sim}, \bibinfo{person}{Shruti Tople}, \bibinfo{person}{Lukas Wutschitz}, {and} \bibinfo{person}{Santiago~Zanella B{\'{e}}guelin}.} \bibinfo{year}{2023}\natexlab{}.
\newblock \showarticletitle{Analyzing Leakage of Personally Identifiable Information in Language Models}. In \bibinfo{booktitle}{\emph{Proceedings of the 44th {IEEE} Symposium on Security and Privacy}}. \bibinfo{publisher}{{IEEE}}, \bibinfo{address}{{San Francisco}, {CA}, {USA}}, \bibinfo{pages}{346--363}.
\newblock
\urldef\tempurl%
\url{https://doi.org/10.1109/SP46215.2023.10179300}
\showDOI{\tempurl}


\bibitem[Mai et~al\mbox{.}(2024)]%
        {heiko-2024}
\bibfield{author}{\bibinfo{person}{Huu~Tan Mai}, \bibinfo{person}{Cuong~Xuan Chu}, {and} \bibinfo{person}{Heiko Paulheim}.} \bibinfo{year}{2024}\natexlab{}.
\newblock \bibinfo{title}{Do LLMs Really Adapt to Domains? An Ontology Learning Perspective}.
\newblock
\newblock
\showeprint[arxiv]{2407.19998}~[cs.CL]
\urldef\tempurl%
\url{https://arxiv.org/abs/2407.19998}
\showURL{%
\tempurl}


\bibitem[Mungall et~al\mbox{.}(2012)]%
        {uberon}
\bibfield{author}{\bibinfo{person}{Christopher~J. Mungall}, \bibinfo{person}{Carlo Torniai}, \bibinfo{person}{Georgios~V. Gkoutos}, \bibinfo{person}{Suzanna~E. Lewis}, {and} \bibinfo{person}{Melissa~A. Haendel}.} \bibinfo{year}{2012}\natexlab{}.
\newblock \showarticletitle{Uberon, an integrative multi-species anatomy ontology}.
\newblock \bibinfo{journal}{\emph{GenomeBiology.com}} \bibinfo{volume}{13}, \bibinfo{number}{1} (\bibinfo{date}{1} \bibinfo{year}{2012}), \bibinfo{numpages}{20}~pages.
\newblock
\showISSN{1465-6906}
\urldef\tempurl%
\url{https://doi.org/10.1186/gb-2012-13-1-r5}
\showDOI{\tempurl}


\bibitem[Mündler et~al\mbox{.}(2024)]%
        {hallucination2}
\bibfield{author}{\bibinfo{person}{Niels Mündler}, \bibinfo{person}{Jingxuan He}, \bibinfo{person}{Slobodan Jenko}, {and} \bibinfo{person}{Martin Vechev}.} \bibinfo{year}{2024}\natexlab{}.
\newblock \bibinfo{title}{Self-contradictory Hallucinations of Large Language Models: Evaluation, Detection and Mitigation}.
\newblock
\newblock
\showeprint[arxiv]{2305.15852}~[cs.CL]
\urldef\tempurl%
\url{https://arxiv.org/abs/2305.15852}
\showURL{%
\tempurl}


\bibitem[Organization(2019)]%
        {ICD-10}
\bibfield{author}{\bibinfo{person}{World~Health Organization}.} \bibinfo{year}{2019}\natexlab{}.
\newblock \bibinfo{title}{International Classification of Diseases, 10th Revision (ICD-10)}.
\newblock
\newblock
\urldef\tempurl%
\url{https://icd.who.int/browse10}
\showURL{%
\tempurl}
\newblock
\shownote{Accessed: October 15, 2024}.


\bibitem[Ranaldi et~al\mbox{.}(2023)]%
        {Zanzotto2023}
\bibfield{author}{\bibinfo{person}{Leonardo Ranaldi}, \bibinfo{person}{Elena~Sofia Ruzzetti}, {and} \bibinfo{person}{Fabio~Massimo Zanzotto}.} \bibinfo{year}{2023}\natexlab{}.
\newblock \showarticletitle{{PreCog}: Exploring the Relation between Memorization and Performance in Pre-trained Language Models}. In \bibinfo{booktitle}{\emph{Proceedings of the 14th International Conference on Recent Advances in Natural Language Processing, 4-6 September 2023}}. \bibinfo{publisher}{{INCOMA} Ltd., Shoumen, Bulgaria}, \bibinfo{address}{Varna, Bulgaria}, \bibinfo{pages}{961--967}.
\newblock


\bibitem[Sainz et~al\mbox{.}(2023)]%
        {sainz-etal-2023-nlp}
\bibfield{author}{\bibinfo{person}{Oscar Sainz}, \bibinfo{person}{Jon Campos}, \bibinfo{person}{Iker Garc{\'\i}a-Ferrero}, \bibinfo{person}{Julen Etxaniz}, \bibinfo{person}{Oier~Lopez de Lacalle}, {and} \bibinfo{person}{Eneko Agirre}.} \bibinfo{year}{2023}\natexlab{}.
\newblock \showarticletitle{{NLP} Evaluation in trouble: On the Need to Measure {LLM} Data Contamination for each Benchmark}. In \bibinfo{booktitle}{\emph{Findings of the Association for Computational Linguistics: EMNLP}}. \bibinfo{publisher}{Association for Computational Linguistics}, \bibinfo{address}{Singapore}, \bibinfo{pages}{10776--10787}.
\newblock
\urldef\tempurl%
\url{https://doi.org/10.18653/v1/2023.findings-emnlp.722}
\showDOI{\tempurl}


\bibitem[Shi et~al\mbox{.}(2024)]%
        {bib-pretraining-detection2-2024}
\bibfield{author}{\bibinfo{person}{Weijia Shi}, \bibinfo{person}{Anirudh Ajith}, \bibinfo{person}{Mengzhou Xia}, \bibinfo{person}{Yangsibo Huang}, \bibinfo{person}{Daogao Liu}, \bibinfo{person}{Terra Blevins}, \bibinfo{person}{Danqi Chen}, {and} \bibinfo{person}{Luke Zettlemoyer}.} \bibinfo{year}{2024}\natexlab{}.
\newblock \bibinfo{title}{Detecting Pretraining Data from Large Language Models}.
\newblock
\newblock
\showeprint[arxiv]{2310.16789}~[cs.CL]
\urldef\tempurl%
\url{https://arxiv.org/abs/2310.16789}
\showURL{%
\tempurl}


\bibitem[Spearman(1904)]%
        {spearman04}
\bibfield{author}{\bibinfo{person}{Charles Spearman}.} \bibinfo{year}{1904}\natexlab{}.
\newblock \showarticletitle{The Proof and Measurement of Association Between Two Things}.
\newblock \bibinfo{journal}{\emph{The American Journal of Psychology}} \bibinfo{volume}{15}, \bibinfo{number}{1} (\bibinfo{year}{1904}), \bibinfo{pages}{72--101}.
\newblock
\showISSN{00029556}
\urldef\tempurl%
\url{http://www.jstor.org/stable/1412159}
\showURL{%
\tempurl}


\bibitem[Sun et~al\mbox{.}(2024)]%
        {bib-knowleadgeable-llms-kg-2024}
\bibfield{author}{\bibinfo{person}{Kai Sun}, \bibinfo{person}{Yifan~Ethan Xu}, \bibinfo{person}{Hanwen Zha}, \bibinfo{person}{Yue Liu}, {and} \bibinfo{person}{Xin~Luna Dong}.} \bibinfo{year}{2024}\natexlab{}.
\newblock \showarticletitle{Head-to-Tail: How Knowledgeable are Large Language Models (LLMs)? {A.K.A.} Will LLMs Replace Knowledge Graphs?}. In \bibinfo{booktitle}{\emph{Proceedings of the 2024 Conference of the North American Chapter of the Association for Computational Linguistics: Human Language Technologies (Volume 1: Long Papers)}}. \bibinfo{publisher}{Association for Computational Linguistics}, \bibinfo{address}{Mexico City, Mexico}, \bibinfo{pages}{311--325}.
\newblock
\urldef\tempurl%
\url{https://doi.org/10.18653/V1/2024.NAACL-LONG.18}
\showDOI{\tempurl}


\bibitem[Tirumala et~al\mbox{.}(2022)]%
        {exact-memorization}
\bibfield{author}{\bibinfo{person}{Kushal Tirumala}, \bibinfo{person}{Aram~H. Markosyan}, \bibinfo{person}{Luke Zettlemoyer}, {and} \bibinfo{person}{Armen Aghajanyan}.} \bibinfo{year}{2022}\natexlab{}.
\newblock \showarticletitle{Memorization Without Overfitting: Analyzing the Training Dynamics of Large Language Models}. In \bibinfo{booktitle}{\emph{Advances in Neural Information Processing Systems}}, Vol.~\bibinfo{volume}{35}. \bibinfo{publisher}{Curran Associates, Inc.}, \bibinfo{address}{New Orleans, LA, USA}, \bibinfo{pages}{38274--38290}.
\newblock
\urldef\tempurl%
\url{https://proceedings.neurips.cc/paper_files/paper/2022/file/fa0509f4dab6807e2cb465715bf2d249-Paper-Conference.pdf}
\showURL{%
\tempurl}


\bibitem[Touvron et~al\mbox{.}(2023)]%
        {touvron2023llama}
\bibfield{author}{\bibinfo{person}{Hugo Touvron}, \bibinfo{person}{Thibaut Lavril}, \bibinfo{person}{Gautier Izacard}, \bibinfo{person}{Xavier Martinet}, \bibinfo{person}{Marie-Anne Lachaux}, \bibinfo{person}{Timothée Lacroix}, \bibinfo{person}{Baptiste Rozière}, \bibinfo{person}{Naman Goyal}, \bibinfo{person}{Eric Hambro}, \bibinfo{person}{Faisal Azhar}, \bibinfo{person}{Aurelien Rodriguez}, \bibinfo{person}{Armand Joulin}, \bibinfo{person}{Edouard Grave}, {and} \bibinfo{person}{Guillaume Lample}.} \bibinfo{year}{2023}\natexlab{}.
\newblock \bibinfo{title}{LLaMA: Open and Efficient Foundation Language Models}.
\newblock
\newblock
\showeprint[arxiv]{2302.13971}~[cs.CL]
\urldef\tempurl%
\url{https://arxiv.org/abs/2302.13971}
\showURL{%
\tempurl}


\bibitem[Vrande\v{c}i\'{c} and Kr\"{o}tzsch(2014)]%
        {wikidata}
\bibfield{author}{\bibinfo{person}{Denny Vrande\v{c}i\'{c}} {and} \bibinfo{person}{Markus Kr\"{o}tzsch}.} \bibinfo{year}{2014}\natexlab{}.
\newblock \showarticletitle{Wikidata: A Free Collaborative Knowledgebase}.
\newblock \bibinfo{journal}{\emph{Commun. ACM}} \bibinfo{volume}{57}, \bibinfo{number}{10} (\bibinfo{date}{sep} \bibinfo{year}{2014}), \bibinfo{pages}{78–85}.
\newblock
\showISSN{0001-0782}
\urldef\tempurl%
\url{https://doi.org/10.1145/2629489}
\showDOI{\tempurl}


\bibitem[Wu et~al\mbox{.}(2023)]%
        {bib:-PLMs-understand-onto2023}
\bibfield{author}{\bibinfo{person}{Weiqi Wu}, \bibinfo{person}{Chengyue Jiang}, \bibinfo{person}{Yong Jiang}, \bibinfo{person}{Pengjun Xie}, {and} \bibinfo{person}{Kewei Tu}.} \bibinfo{year}{2023}\natexlab{}.
\newblock \showarticletitle{Do PLMs Know and Understand Ontological Knowledge?}. In \bibinfo{booktitle}{\emph{Proceedings of the 61st Annual Meeting of the Association for Computational Linguistics (Volume 1: Long Papers)}}. \bibinfo{publisher}{Association for Computational Linguistics}, \bibinfo{address}{Toronto, Canada}, \bibinfo{pages}{3080--3101}.
\newblock
\urldef\tempurl%
\url{https://doi.org/10.18653/v1/2023.acl-long.173}
\showDOI{\tempurl}


\bibitem[Zhang et~al\mbox{.}(2023)]%
        {hallucination0}
\bibfield{author}{\bibinfo{person}{Yue Zhang}, \bibinfo{person}{Yafu Li}, \bibinfo{person}{Leyang Cui}, \bibinfo{person}{Deng Cai}, \bibinfo{person}{Lemao Liu}, \bibinfo{person}{Tingchen Fu}, \bibinfo{person}{Xinting Huang}, \bibinfo{person}{Enbo Zhao}, \bibinfo{person}{Yu Zhang}, \bibinfo{person}{Yulong Chen}, \bibinfo{person}{Longyue Wang}, \bibinfo{person}{Anh~Tuan Luu}, \bibinfo{person}{Wei Bi}, \bibinfo{person}{Freda Shi}, {and} \bibinfo{person}{Shuming Shi}.} \bibinfo{year}{2023}\natexlab{}.
\newblock \bibinfo{title}{Siren's Song in the AI Ocean: A Survey on Hallucination in Large Language Models}.
\newblock
\newblock
\showeprint[arxiv]{2309.01219}~[cs.CL]
\urldef\tempurl%
\url{https://arxiv.org/abs/2309.01219}
\showURL{%
\tempurl}


\bibitem[Zhou et~al\mbox{.}(2024a)]%
        {bib-black-box-llm-2024}
\bibfield{author}{\bibinfo{person}{Baohang Zhou}, \bibinfo{person}{Zezhong Wang}, \bibinfo{person}{Lingzhi Wang}, \bibinfo{person}{Hongru Wang}, \bibinfo{person}{Ying Zhang}, \bibinfo{person}{Kehui Song}, \bibinfo{person}{Xuhui Sui}, {and} \bibinfo{person}{Kam{-}Fai Wong}.} \bibinfo{year}{2024}\natexlab{a}.
\newblock \showarticletitle{{DPDLLM:} {A} Black-box Framework for Detecting Pre-training Data from Large Language Models}. In \bibinfo{booktitle}{\emph{Findings of the Association for Computational Linguistics: {ACL}}}. \bibinfo{publisher}{Association for Computational Linguistics}, \bibinfo{address}{Bangkok, Thailand and virtual meeting}, \bibinfo{pages}{644--653}.
\newblock
\urldef\tempurl%
\url{https://doi.org/10.18653/v1/2024.findings-acl.35}
\showDOI{\tempurl}


\bibitem[Zhou et~al\mbox{.}(2024b)]%
        {bib-entity-level-memorization-2024}
\bibfield{author}{\bibinfo{person}{Zhenhong Zhou}, \bibinfo{person}{Jiuyang Xiang}, \bibinfo{person}{Chaomeng Chen}, {and} \bibinfo{person}{Sen Su}.} \bibinfo{year}{2024}\natexlab{b}.
\newblock \showarticletitle{Quantifying and Analyzing Entity-Level Memorization in Large Language Models}. In \bibinfo{booktitle}{\emph{Proceedings of the Thirty-Eighth AAAI Conference on Artificial Intelligence and Thirty-Sixth Conference on Innovative Applications of Artificial Intelligence and Fourteenth Symposium on Educational Advances in Artificial Intelligence}}. \bibinfo{publisher}{{AAAI} Press}, \bibinfo{address}{Vancouver, Canada}, \bibinfo{pages}{19741--19749}.
\newblock
\urldef\tempurl%
\url{https://doi.org/10.1609/aaai.v38i17.29948}
\showDOI{\tempurl}


\bibitem[Zhu et~al\mbox{.}(2024)]%
        {hallucination1}
\bibfield{author}{\bibinfo{person}{Derui Zhu}, \bibinfo{person}{Dingfan Chen}, \bibinfo{person}{Qing Li}, \bibinfo{person}{Zongxiong Chen}, \bibinfo{person}{Lei Ma}, \bibinfo{person}{Jens Grossklags}, {and} \bibinfo{person}{Mario Fritz}.} \bibinfo{year}{2024}\natexlab{}.
\newblock \showarticletitle{PoLLMgraph: Unraveling Hallucinations in Large Language Models via State Transition Dynamics}. In \bibinfo{booktitle}{\emph{Findings of the Association for Computational Linguistics: {NAACL}}}. \bibinfo{publisher}{Association for Computational Linguistics}, \bibinfo{address}{Mexico City, Mexico}, \bibinfo{pages}{4737--4751}.
\newblock
\showISBNx{979-8-89176-119-3}
\urldef\tempurl%
\url{https://doi.org/10.18653/v1/2024.findings-naacl.294}
\showDOI{\tempurl}


\end{thebibliography}
\end{document}